\pdfoutput=1

\documentclass[11pt]{article}

\usepackage[final]{acl}

\usepackage{times}
\usepackage{latexsym}
\usepackage[T1]{fontenc}
\usepackage[utf8]{inputenc}
\usepackage{microtype}
\usepackage{inconsolata}
\usepackage{graphicx}
\usepackage{enumitem}
\usepackage{xspace}
\usepackage{amsmath}
\usepackage{booktabs}

\usepackage{multirow}
\usepackage{multicol}
\usepackage{pifont}

\newcommand{\method}{\textsc{GLN}\xspace}

\definecolor{best}{RGB}{120,220,190}

\newcommand{\A}{%
  \textcolor{red}{\textcircled{\textcolor{red}{\small A}}}%
}

\definecolor{lightgreen}{rgb}{0.88, 1, 0.88}
\newcommand{\cmark}{\textcolor{green!60!black}{\ding{51}}} 
\newcommand{\xmark}{\textcolor{red}{\ding{55}}}            

\title{`Hello, World!': Making GNNs Talk with LLMs}

\author{Sunwoo Kim\textsuperscript{1} \quad\quad
        Soo Yong Lee\textsuperscript{1} \quad\quad
        Jaemin Yoo\textsuperscript{2} \quad\quad 
        Kijung Shin\textsuperscript{1,2} \\
        \textsuperscript{1}Kim Jaechul Graduate School of AI, KAIST, \quad  \textsuperscript{2}School of Electrical Engineering, KAIST \\
        \texttt{\{kswoo97, syleetolow, jaemin, kijungs\}@kaist.ac.kr}}

\begin{document}
\maketitle
\begin{abstract}

While graph neural networks (GNNs) have shown remarkable performance across diverse graph-related tasks, their high-dimensional hidden representations render them black boxes.
In this work, we propose Graph Lingual Network (GLN), a GNN built on large language models (LLMs), with hidden representations in the form of human-readable text.
Through careful prompt design, GLN incorporates not only the message passing module of GNNs but also advanced GNN techniques, including graph attention and initial residual connection.
The comprehensibility of GLN’s hidden representations enables an intuitive analysis of how node representations change (1) across layers and (2) under advanced GNN techniques, shedding light on the inner workings of GNNs.
Furthermore, we demonstrate that GLN achieves strong zero-shot performance on node classification and link prediction, outperforming existing LLM-based baseline methods.
\end{abstract}

\section{Introduction}
\label{sec:introduction}

Graph neural networks (GNNs) are designed to process graph-structured data, and they have demonstrated strong performance in various downstream tasks such as node classification and link prediction~\cite{corso2024graph}.
A key to their success lies in their message passing module, which updates the representation of a node by aggregating information from its neighbors~\cite{hamilton2020graph}.
However, the high-dimensional embeddings (i.e., vectorized representations) obtained via existing GNNs are generally not comprehensible.


In this work, we propose \textbf{GLN} ({\underline{\textbf{G}}raph \underline{\textbf{L}}ingual \underline{\textbf{N}}etwork}), where an LLM is prompted to aggregate neighbor information to update a node's representation.
Therefore, all hidden node representations of \method are human-readable texts.
Moreover, we propose a tailored LLM prompting framework incorporating advanced GNN techniques, specifically graph attention~\cite{velivckovic2017graph} and initial residual connection~\cite{chen2020simple}.

Compared to existing GNNs, our \method offers several advantages.
First, its hidden representations are \textit{comprehensible and human-readable}, since they are text descriptions of nodes generated by the LLM.
Second, using an LLM as the message passing module enables GLN to solve graph-related tasks in a zero-shot manner, without any training or task labels.
Third, \method can be further prompted to \textit{explain its decisions} on graph-related tasks, facilitating human understanding of its reasoning.

Thanks to the comprehensibility of \method's hidden representations, we provide an intuitive analysis regarding how the node representations change (1) across GLN layers and (2) under advanced GNN techniques.
Drawing from this analysis, we offer several key insights into the mechanisms underlying GNN message passing and its advanced techniques.
Moreover, we demonstrate the zero-shot capability of \method on popular downstream tasks (node classification and link prediction), demonstrating its superiority over existing LLM-based baseline methods.
Code, datasets, and example node representations generated by \method are in~\url{https://github.com/kswoo97/GLN-Code}.




\section{Related work and preliminaries}
\label{sec:relatedwork}
In this section, we cover related work and preliminaries of our research.

\begin{figure*}[t!]
    \centering
    
    \includegraphics[width=1.0\textwidth]{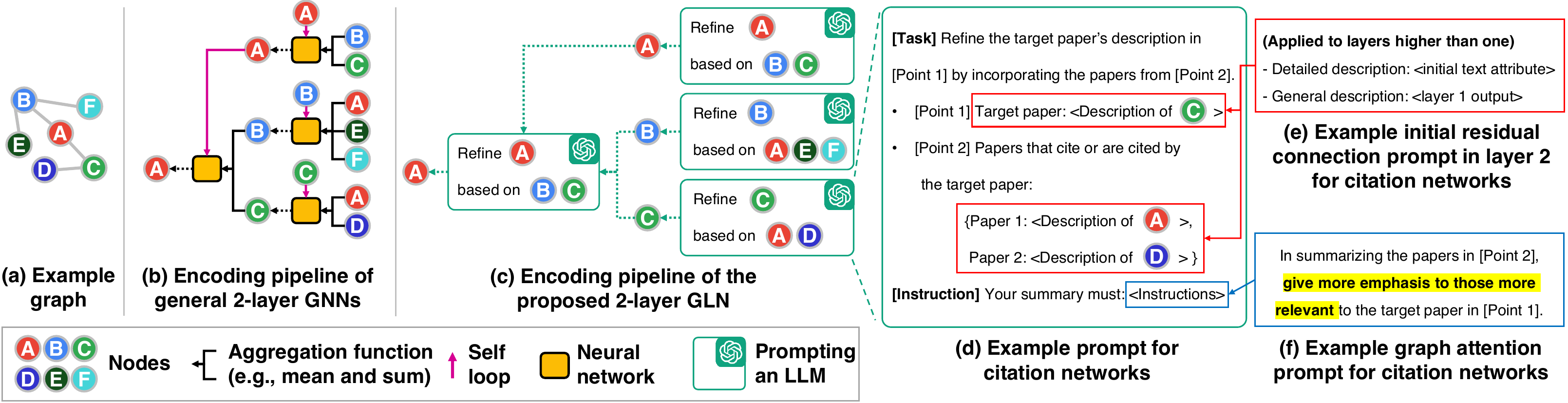}
    \vspace{-5mm}
    \caption{\textbf{Overview of general GNNs and our \method.}
    To obtain node \textbf{\A}’s representation in example graph (a), a general 2-layer GNN first refines the representations of \textbf{\A} and its one-hop neighbors, and then updates that of \textbf{\A} again using these refined representations, as shown in (b).
    A similar mechanism is applied in \method, as shown in (c), but the aggregation functions and neural networks are replaced by an LLM with our prompt, illustrated in (d) - (f).}   
    \label{fig:method}
    \vspace{-5mm}
\end{figure*}

\subsection{Related work}\label{subsec:relatedwork}

\noindent\textbf{Graph neural networks.}
In various graph-related tasks, such as node classification and link prediction, graph neural networks (GNNs) have achieved strong performance~\cite{luo2024classic}.
The core module of a GNN involves message-passing, which updates node representation by aggregating information from its neighboring nodes~\cite{hamilton2020graph} (refer to Figure~\ref{fig:method} (b) for an example).
This process is typically repeated across layers, enabling a node's representation at $k-$th layer to summarize its $k-$hop neighbor information.

Several advanced techniques have been proposed to enhance GNN message passing, notably \textit{graph attention}~\cite{velivckovic2017graph} and \textit{initial residual connection}~\cite{chen2020simple}.
Graph attention allows a GNN to learn the relative importance of each neighbor during aggregation. 
Initial residual connections help preserve original representations (i.e., initial feature vectors) by injecting them into each layer, mitigating their degradation by the repeated message passing.

\noindent\textbf{Combining GNNs with LLMs.}
With the remarkable performance of LLMs in a wide range of domains~\cite{chang2024survey}, many studies have combined them with GNNs to tackle various graph-related tasks~\cite{ren2024survey}.
Early works either fine-tuned LLMs~\cite{chen2024llaga} or fed LLM outputs into GNNs during their training for graph-related tasks~\cite{he2023harnessing}, both incurring high training costs.
In contrast, several recent works prompted LLMs to model GNNs' operations without further training~\cite{chen2024exploring, zhu2025llm}. 
Among them, \citet{zhu2025llm} obtained graph vocabulary for graph foundation models by prompting LLMs to mimic the message-passing modules of GNNs.
Since their method does not aim for user comprehension, the refined representations offer limited utility from a comprehension perspective.
Specifically, instead of enriching textual representations of nodes across layers, it tends to merely enumerate neighbor information (see Appendix~\ref{subapp:promptgfmours} for further details).


\subsection{Preliminaries}\label{subsec:prelim}
A graph $\mathcal{G}=\{\mathcal{V}, \mathcal{E}\}$ is defined by a node set $\mathcal{V}=\{v_{1},\cdots,v_{\vert \mathcal{V}\vert}\}$ and an edge set $\mathcal{E}=\{e_{1},\cdots,e_{\vert \mathcal{E}\vert}\}$.
Each edge $e_{i} \in \mathcal{E}$ is defined by a pair of nodes (i.e., $e_{i} \in \binom{\mathcal{V}}{2}$), and node $v_{i}$'s neighbor set $\mathcal{N}_{i}$ is defined by a set of nodes linked to $v_{i}$ (i.e., $\mathcal{N}_{i} = \{v_{j} \in \mathcal{V} : \{v_{i},v_{j}\} \in \mathcal{E}\}$).
In this work, we consider a \textit{text-attributed graph}, where each node $v_{i} \in \mathcal{V}$ is associated with a text attribute ${D}^{(0)}_{i}$ that describes $v_{i}$, which we call initial text attribute.


\section{Proposed method: \method}
\label{sec:method}
\begin{figure*}[t!]
    \centering
    {
    \includegraphics[width=1.0\textwidth]{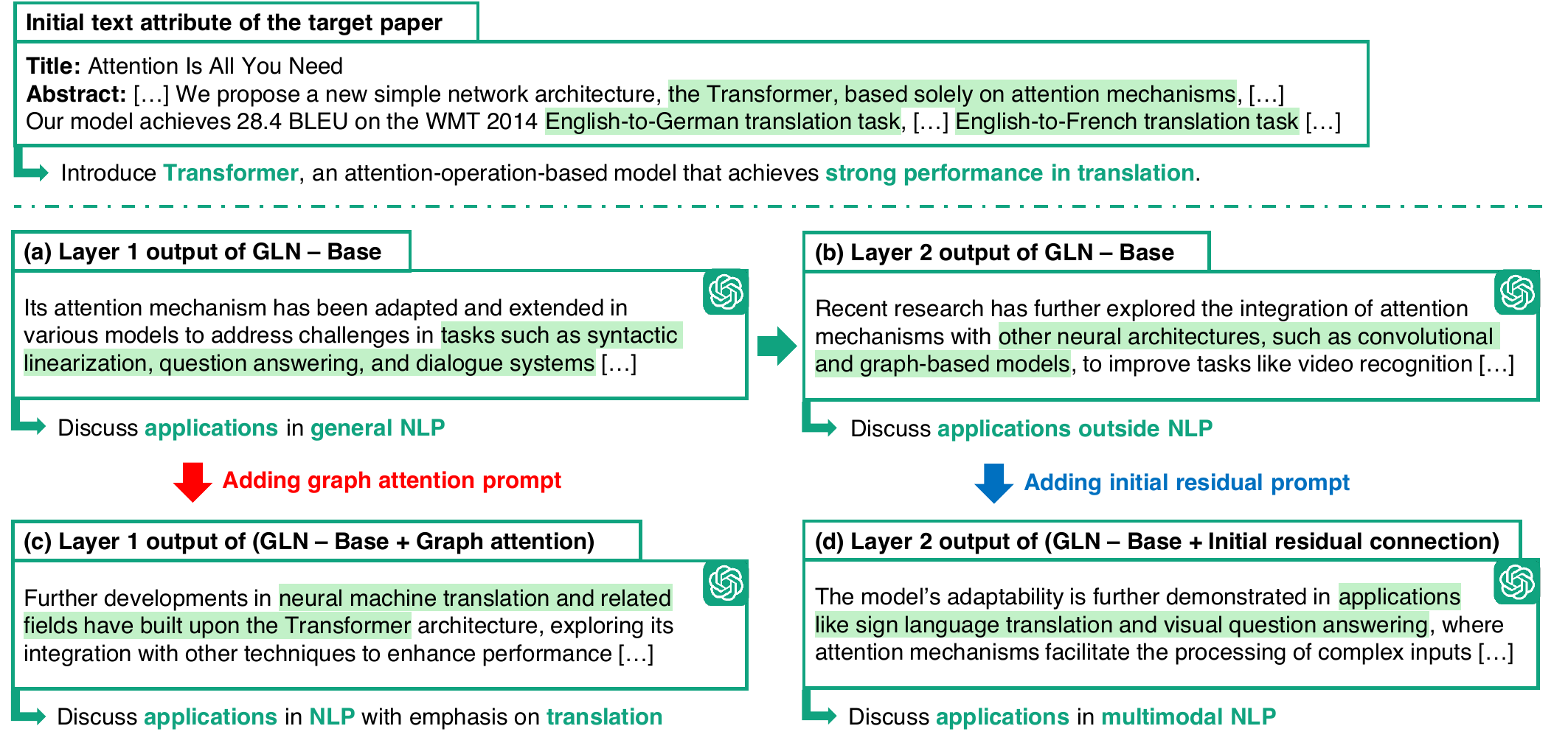}
    }
    \vspace{-5mm}
    \caption{\textbf{Representations become more general across layers, are specialized to the target node through graph attention, and retain target-specific information in higher layers through initial residual connections.} 
    We present a case study on~\citep{vaswani2017attention}, showing \method-Base's representations of the paper at: (a) layer 1, (b) layer 2, (c) layer 1 with a graph attention prompt, and (d) layer 2 with an initial residual connection prompt.}  
    \label{fig:casestudy}  
    \vspace{-5mm}
\end{figure*}

In this section, we introduce \textbf{\method} ({\underline{\textbf{G}}raph \underline{\textbf{L}}ingual \underline{\textbf{N}}etwork}), a graph neural network where an LLM serves as its message passing module.
We first give an overview of \method (Sec.~\ref{subsec:methodoverview}) and describe our specialized prompt that incorporates GNNs' advanced techniques (Sec.~\ref{subsec:methodprompt}). 
Refer to Figure~\ref{fig:method} for an overview of \method.

\subsection{Overview}\label{subsec:methodoverview}

At each layer, \method refines the textual representation of each node by prompting an LLM to aggregate information from the node's neighbors. 
Specifically, at layer $\ell$, an LLM receives an input prompt consisting of (1) the target node $v_{i}$'s representation from the previous layer $D^{(\ell-1)}_{i}$ and (2) the neighbor representations from the previous layer $\{D^{(\ell-1)}_{j}:v_{j}\in \mathcal{N}_{i}\}$ (the prompt design is detailed in Sec.~\ref{subsec:methodprompt}).~\footnote{Recall that $D_{i}^{(0)}$ is the $v_{i}$'s initial text attribute (Sec.~\ref{subsec:prelim}).}
The LLM then outputs the refined textual representation of $v_{i}$, denoted as $D^{(\ell)}_{i}$, effectively integrating the prior representations of both the target node and its neighbors. 

After $L$ iterations, corresponding to the number of \method layers, we define the final representation of node $v_{i}$ as the set of its intermediate embeddings (i.e., $\{D^{(t)}_{i}:t\in \{0, 1,\cdots, L\}$) \footnote{Detailed representation format is in Appendix~\ref{subapp:representationformat}.}
to capture diverse information about the node. 
Here, each layer of a GNN captures a different level of information: the earlier layers encode fine-grained, local features from immediate neighbors, while the later layers aggregate broader and more abstract information from multi-hop neighbors~\cite{xu2018representation}.

\vspace{-2mm}
\subsection{Advanced techniques}\label{subsec:methodprompt}
\textbf{GNN-style prompting.}
The key innovation of \method involves its prompt design, which determines how the LLM aggregates neighbor information to update the target node representation. 
For this, we adopt two advanced GNN techniques: (1) {graph attention}~\cite{velivckovic2017graph} and (2) {initial residual connection}~\cite{chen2020simple} (refer to Sec.~\ref{subsec:relatedwork} for their details).
To implement graph attention with an LLM, we design a prompt that encourages the LLM to place greater emphasis on neighbors that are more relevant to the target node during aggregation; we refer to this as the \textit{graph attention prompt} (refer to Figure~\ref{fig:method} (f)).~\footnote{We provide further analysis on the effect of the graph attention prompt in Appendix~\ref{subapp:graphattention}.}
Similarly, to implement initial residual connections, we include both the previous-layer output and the raw attributes of each node in their descriptions; we refer to this as the \textit{initial residual connection prompt} (refer to Figure~\ref{fig:method} (e)). 
Details on our prompt design and its alternatives are provided in Appendix~\ref{subapp:glnencoding}.

\noindent\textbf{Token-efficient prompting.}
We can further improve the efficiency of \method by incorporating a token-efficient prompting strategy that reduces input and/or output tokens.
Input tokens can be reduced by updating node representations with randomly sampled neighbors rather than the full neighborhood.
Moreover, output tokens can be reduced by instructing the LLM to follow a specific format, such as limiting the number of generated paragraphs.
In our experiments, we fix the number of neighbor samples at 10 and limit the output to 2 paragraphs. 
Nevertheless, we validate that \method remains competitive 
even when fewer neighbor samples are used and when it is prompted to produce shorter outputs, 
as detailed in Appendix~\ref{subapp:scalability}.



\section{Analysis and experiments}
\label{sec:analysis}

\begin{table*}[t]
\centering
\small

\setlength{\tabcolsep}{3.0pt} 
\renewcommand{\arraystretch}{1.0}

\begin{tabular}{c|l| ccc | ccc | c}
\toprule
 \multirow{2}{*}{LLMs} & \multirow{2}{*}{Methods} & \multicolumn{3}{c|}{Task: Node classification} & \multicolumn{3}{c|}{Task: Link prediction} & \multirow{2}{*}{A.R.}\\

 & & OGBN-Arxiv & Book-History & Wiki-CS & OGBN-Arxiv & Book-History & Wiki-CS  & \\

\midrule
\midrule

\multirow{5}{*}{\rotatebox[origin=c]{90}{GPT}} & \texttt{Direct} & 62.3 & 44.7 & 78.3 & 92.8 & 85.0 & 83.2 & 3.8\\

 & \texttt{All-in-One} & 61.5 & 45.4 & 64.9 & 91.6 & 84.8 & \cellcolor{best}\textbf{85.6} & 3.6 \\

 & \texttt{PromptGFM} & 62.0 & 44.1 & 79.0 & 92.2 & 81.2 & 81.0 & 4.2 \\

\cmidrule{2-9}

 & \method-Base & 63.0 & 45.8 & 79.4 & 92.4 & 86.4 & 83.6 & 2.2 \\
 & \method & \cellcolor{best}\textbf{64.0} & \cellcolor{best}\textbf{47.3} & \cellcolor{best}\textbf{79.5} & \cellcolor{best}\textbf{93.0} & \cellcolor{best}\textbf{87.0} & 84.0 & \cellcolor{best}\textbf{1.2}\\

 \midrule
 \midrule

 \multirow{5}{*}{\rotatebox[origin=c]{90}{Claude}} & \texttt{Direct} & 65.8 & 48.4 & 76.6 & 78.0 & \cellcolor{best}\textbf{65.2} & 41.2 & 3.2 \\

 & \texttt{All-in-One} & 67.1 & 50.4 & 76.4 & 72.8 & 53.6 & 38.6 & 3.7 \\

 & \texttt{PromptGFM} & 65.3 & 50.6 & 74.5 & 64.4 & 50.0 & 38.2 & 4.7 \\

\cmidrule{2-9}

 & \method-Base & 67.1 & 53.8 & 77.0 & 78.2 & 61.2 & 42.4 & 2.2 \\
 
 & \method & \cellcolor{best}\textbf{67.4} & \cellcolor{best}\textbf{55.2} & \cellcolor{best}\textbf{77.7} & \cellcolor{best}\textbf{78.4} & 64.0 & \cellcolor{best}\textbf{43.2} & \cellcolor{best}\textbf{1.2}\\

\bottomrule
\end{tabular}
\vspace{-2mm}
\caption{\textbf{\method outperforms the zero-shot LLM-based baselines on popular graph-related tasks.}
For node classification and link prediction, we report accuracy and Hit-ratio@1, respectively, of each method in each dataset.
A.R. denotes average ranking.
The best performance in each setting is highlighted in \colorbox{best}{\textbf{green}}.}
\label{tab:mainresult}
\vspace{-6mm}
\end{table*}

In this section, we analyze representations obtained by \method and demonstrate its zero-shot capability in several graph-related tasks.

\subsection{Representation analysis}\label{subsec:qualitative}


\noindent\textbf{Setup.} 
We conduct a case study of \method representation of an academic paper, \citep{vaswani2017attention} (Figure~\ref{fig:casestudy}), on the OGBN-arXiv citation network dataset~\cite{hu2020open}, where nodes and edges represent papers and citations, respectively.
Additional examples from diverse domains (e.g., computer vision and graph learning) are in Appendix~\ref{subapp:representationanalysis}. 
We extract layer-1 and layer-2 outputs of \textbf{\method-Base}, a \method variant that omits graph attention and initial residual connection prompts, using GPT-4o to analyze the effect of the message passing. 
To analyze the impact of the two GNN techniques, we also extract the paper's \method representation with the graph attention prompt and one with the initial residual connection prompt.

\noindent\textbf{\textit{Observation 1. The node representations become more general across layers.}}
As shown in Figure~\ref{fig:casestudy} (a) and (b), the layer-1 representation focuses on how the attention operation is used in the NLP domain.
In the layer-2, the scope expands to applications of attention in computer vision and graph learning.
This shift across \method layers suggests that adding message-passing layers (i.e., incorporating information of the farther neighbors) makes each node’s representation more general.

\noindent\textbf{\textit{Observation 2. Graph attention tailors the neighbor summary for the target node.}}
As shown in Figure~\ref{fig:casestudy} (a) and (c), the paper representation obtained from \method-Base involves non-specific enumeration of various NLP tasks. 
However, after applying the graph attention prompt, the representation involves the specific task addressed in the target paper.
This result suggests that graph attention encourages aggregation toward neighbors that are more relevant to the target paper.

\noindent\textbf{\textit{Observation 3. Initial residual connection preserves the target node information after message passing.}}
As shown in Figure~\ref{fig:casestudy} (c) and (d), the representation obtained from \method-Base describes application of the attention operation outside NLP, whereas the one obtained after applying the initial residual prompt maintains its focus on the NLP domain.
This result suggests that the initial residual connection prompt preserves the target node’s initial text attribute, encouraging its updated representation to stay aligned with that context.

\begin{figure}[t!]
    \centering
    \includegraphics[width=1.0\linewidth]{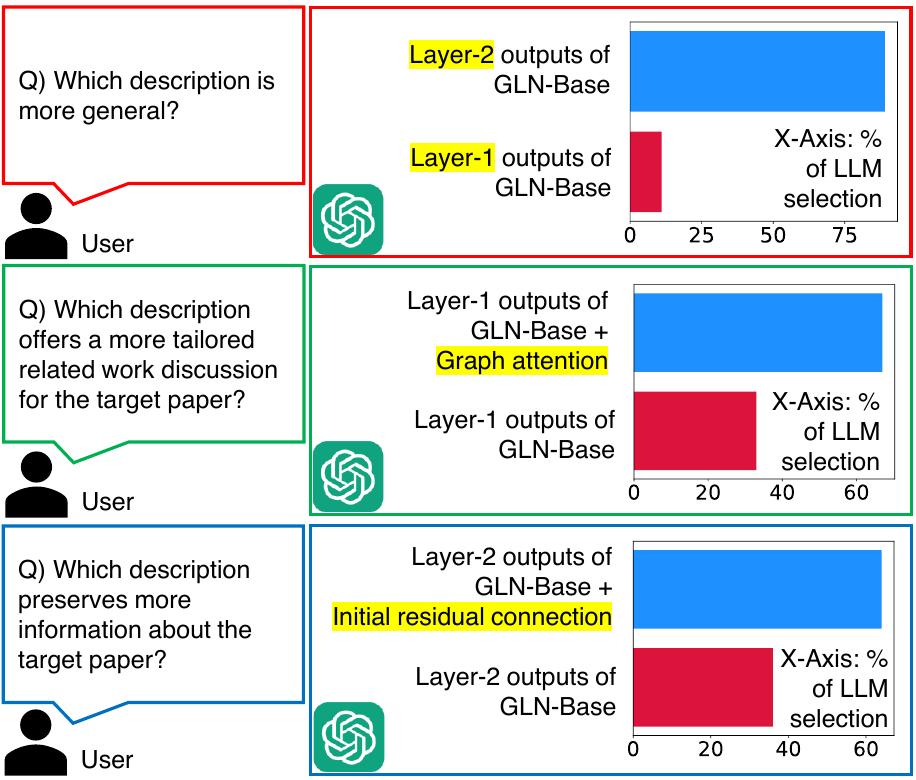}
    \caption{\textbf{\textit{Observations 1-3} hold valid under the LLM-as-a-judge protocol}.
    We report the ratio of LLM responses per category for each question.}  
    \label{fig:llmevaluation}  
    \vspace{-5mm}
\end{figure}

\noindent\textbf{\textit{Observation 4. Observations 1-3 are validated via LLM-as-a-judge protocol.}}
We provide a quantitative assessment of our observations.
To this end, we randomly sample $10^2$ papers and extract the four aforementioned representations for each.
We then prompt GPT-4o to validate our observation on the representations, resulting in an evaluation consistent with our observations, as shown in Figure~\ref{fig:llmevaluation}.

\subsection{Zero-shot capability analysis}\label{subsec:quantitative}

\noindent\textbf{Setup.}
We use two backbone LLMs (GPT-4o-mini and Claude-3.0-Haiku)~\footnote{In Appendix~\ref{subapp:scalability}, we present an analysis with a small language model, showing competitive performance while offering faster inference compared to larger LLMs.} and three real-world graphs: a citation network (OGBN-arXiv), a co-purchase network (Book-History), and a hyperlink network (Wiki-CS), whose further details are in Appendix~\ref{appendix:dataset}.
After obtaining the target node’s textual representation using the proposed method and baselines, we input it into an LLM to perform node classification and edge prediction.
Detailed prompts for each task and further experimental details in Appendices~\ref{subapp:detailtaskprompt} and \ref{subapp:settingdetail}, respectively.


\noindent\textbf{Baseline methods and \method.}
We use four baseline methods:
(1) using only the initial text attribute of the target node (\texttt{Direct}),
(2) providing one-hop and two-hop neighbors to the LLM to update the target node’s representation (\texttt{All-in-One}),
(3) an existing text-attributed graph foundation model (\texttt{PromptGFM})~\cite{zhu2025llm}, and
(4) a baseline version of \method (\method-Base).
For fair comparison, all methods—including the baselines and \method—are equipped with the same backbone LLMs.
Further details regarding baselines and \method are provided in Appendix~\ref{subapp:baselinedetail}.

\noindent\textbf{Results.}
\method outperforms baseline methods in 10 out of 12 settings (Table~\ref{tab:mainresult}), demonstrating its effectiveness in zero-shot capability in node classification and link prediction. 
Two points stand out: (1) \method’s superior performance over \texttt{Direct} highlights the effectiveness of utilizing graph topology, and (2) its gain over \method-Base shows the effectiveness of our {advanced GNN-style} prompts.
Further ablation study results are in Appendix~\ref{subapp:ablation}.

\noindent\textbf{LLM Reasoning.}
We further analyze the LLM’s reasoning behind its downstream task decisions in Appendix~\ref{subapp:llmreasoning}, highlighting which parts of the textual representation contribute to performance.
\vspace{-2mm}

\section{Conclusion}
\label{sec:conclusion}

In this work, we propose \method, a GNN that uses an LLM as its message passing module (Sec.~\ref{sec:method}). 
Leveraging the comprehensibility of its hidden representations, we provide intuitive insights into message passing and advanced GNN techniques (Sec.~\ref{subsec:qualitative}). 
Moreover, \method outperforms baselines on zero-shot graph-related tasks (Sec.~\ref{subsec:quantitative}).







\section*{Limitations}
\noindent\textbf{Theoretical property.} 
Various theoretical properties of GNNs, such as expressivity~\cite{xu2018powerful} and permutation invariance~\cite{keriven2019universal}, have been widely studied. 
Such analyses rely on certain theoretical properties of the neighbor aggregation functions used in GNNs.
However, since \method employs an LLM as its aggregation function, deriving the analogous properties is challenging. 
Thus, the theoretical properties of \method remain underexplored in this work and can be a promising direction for future work.

\vspace{2mm}

\noindent\textbf{Computational efficiency compared to GNNs.}
Due to the usage of a large language model, \method has significantly more parameters than general GNNs, which return vectorized node representations.
Therefore, exploring scaled-up versions of \method can be a promising direction for future work.

\vspace{2mm}

\noindent\textbf{LLM API cost.}
We used LLM APIs (specifically, GPT-4o, GPT-4o-mini, and Claude-3.0-Haiku), incurring a total cost of approximately \$600 for this research.
This may hinder broader accessibility and practical use in budget-constrained environments.
While \method incorporates token-efficient prompting (Sec.~\ref{subsec:methodprompt}), enhancing token-efficiency further may extend the applicability of our approach.

\vspace{2mm}

\noindent\textbf{Extensions to various graph types.}
In this work, we focus on text-attributed graphs (TAGs), where (1) each node is associated with a text attribute and (2) each edge represents a relation between two nodes.
However, many real-world graphs go beyond text attributes or pairwise relations.
Specifically, these include (1) non-text-attributed graphs, such as sensor networks with numerical node features~\citep{jablonski2017graph}, and (2) higher-order relations among multiple nodes, typically modeled as hypergraphs~\citep{kim2024survey, kim2024hypeboy}. 
Thus, extending \method to support such graph types can improve its application to a broader set of real-world scenarios.

\vspace{2mm}

\noindent\textbf{Noisy node text attributes.} 
In this work, we use the TAG benchmark datasets in which node attributes have been carefully preprocessed by their original curators.
However, real-world node text attributes often contain noise (e.g., low-quality reviews in co-purchase networks), which can harm GNN performance~\cite{yan2023comprehensive}.
Our preliminary analysis also shows that \method suffers performance degradation when noise is introduced into the input node attributes (see Appendix~\ref{subsec:noisyinput}).
Thus, incorporating text denoising into \method can improve its practicality in cases with noisy node attributes.

\section*{Acknowledgments}
This work was partly supported by the National Research Foundation of Korea (NRF) grant funded by the Korea government (MSIT) (No. RS-2024-00406985, 30\%).
This work was partly supported by Institute of Information \& Communications Technology Planning \& Evaluation (IITP) grant funded by the Korea government (MSIT)  (No. RS-2024-00457882, AI Research Hub Project, 30\%)
(No. 2022-0-00871 / RS-2022-II220871, Development of AI Autonomy and Knowledge Enhancement for AI Agent Collaboration, 30\%)
(RS-2019-II190075, Artificial Intelligence Graduate School Program (KAIST), 10\%).

\bibliography{custom}

\appendix

\section{Dataset details}
\label{appendix:dataset}
\begin{table}[t] 
\centering
\begin{tabular}{lccc}
\toprule
\textbf{Dataset} & \textbf{\#Nodes} & \textbf{\#Edges} & \textbf{\#Classes} \\
\midrule
OGBN-Arxiv     & 169,343 & 1,166,243 & 40 \\
Book-History & 41,551 & 358,574 & 12 \\
Wiki-CS   & 11,701 & 216,123 & 10 \\
\bottomrule
\end{tabular}
\caption{Graph statistics of the datasets.}
\label{tab:graph_stats}
\end{table}

In this appendix section, we provide details regarding the datasets used in this work.
The detailed statistics of each dataset is provided in Table~\ref{tab:graph_stats}.

\noindent\textbf{OGBN-Arxiv}~\cite{hu2020open} is a citation network that represents the citation relations between papers. 
In this dataset, each node corresponds to a particular paper, and edges join the papers that cite or are cited by the corresponding paper. 
The attributes of a node correspond to the title and abstract of the corresponding node (paper).
The class of a node corresponds to the arXiv category to which the corresponding node (paper) belongs.

\noindent\textbf{Book-History}~\cite{yan2023comprehensive} is a co-purchase network that represents the co-purchase relations among books.
In this dataset, each node corresponds to a particular book, and edges join the books that are frequently co-purchased together with the corresponding book.
The attributes of a node correspond to the description of the corresponding node (book).
The class of a node corresponds to the Amazon third-level category to which the corresponding node (book) belongs.

\noindent\textbf{Wiki-CS}~\cite{mernyei2020wiki} is a hyperlink network that represents the hyperlink relations among Wikipedia web pages.
In this dataset, each node corresponds to a particular Wikipedia web page, and edges join the pages that are either hyperlinked to or from the corresponding page.
The attributes of a node correspond to the content within the corresponding node (web page).
The class of a node corresponds to the Wikipedia category to which the corresponding node (web page) belongs.

\section{Additional analysis}
\label{appendix:casestudy}
\begin{figure*}[t!]
    \centering
    \fbox{
    \includegraphics[width=1.0\textwidth]{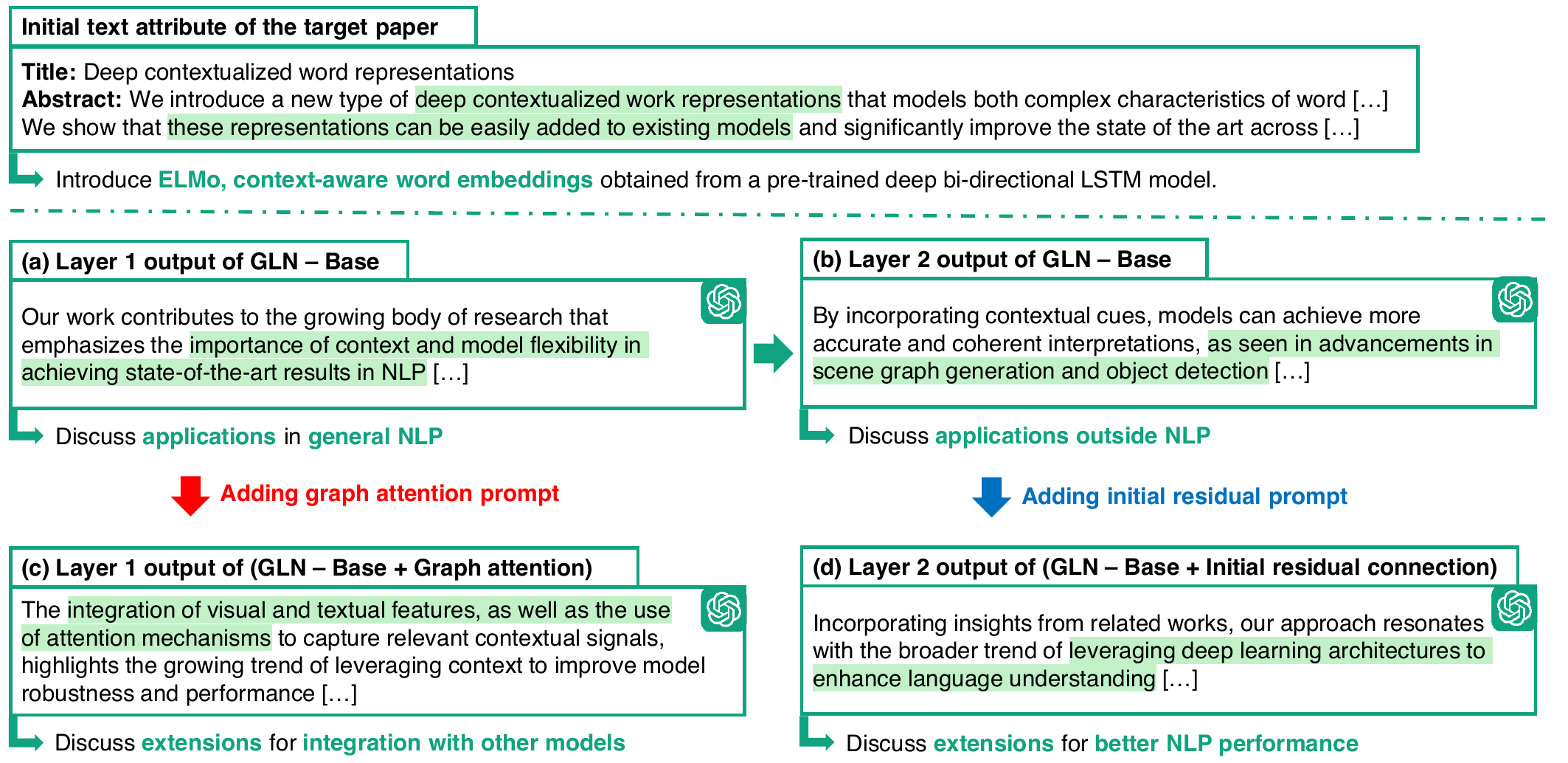}
    }
    \caption{A case study on~\citep{peters2018deep}, showing \method-Base's representations of the paper at: (a) layer 1, (b) layer 2, (c) layer 1 with a graph attention prompt, and (d) layer 2 with an initial residual connection prompt.} 
    \label{fig:elmocasestudy}  
\end{figure*}

\begin{figure*}[t!]
    \centering
    \fbox{
    \includegraphics[width=1.0\textwidth]{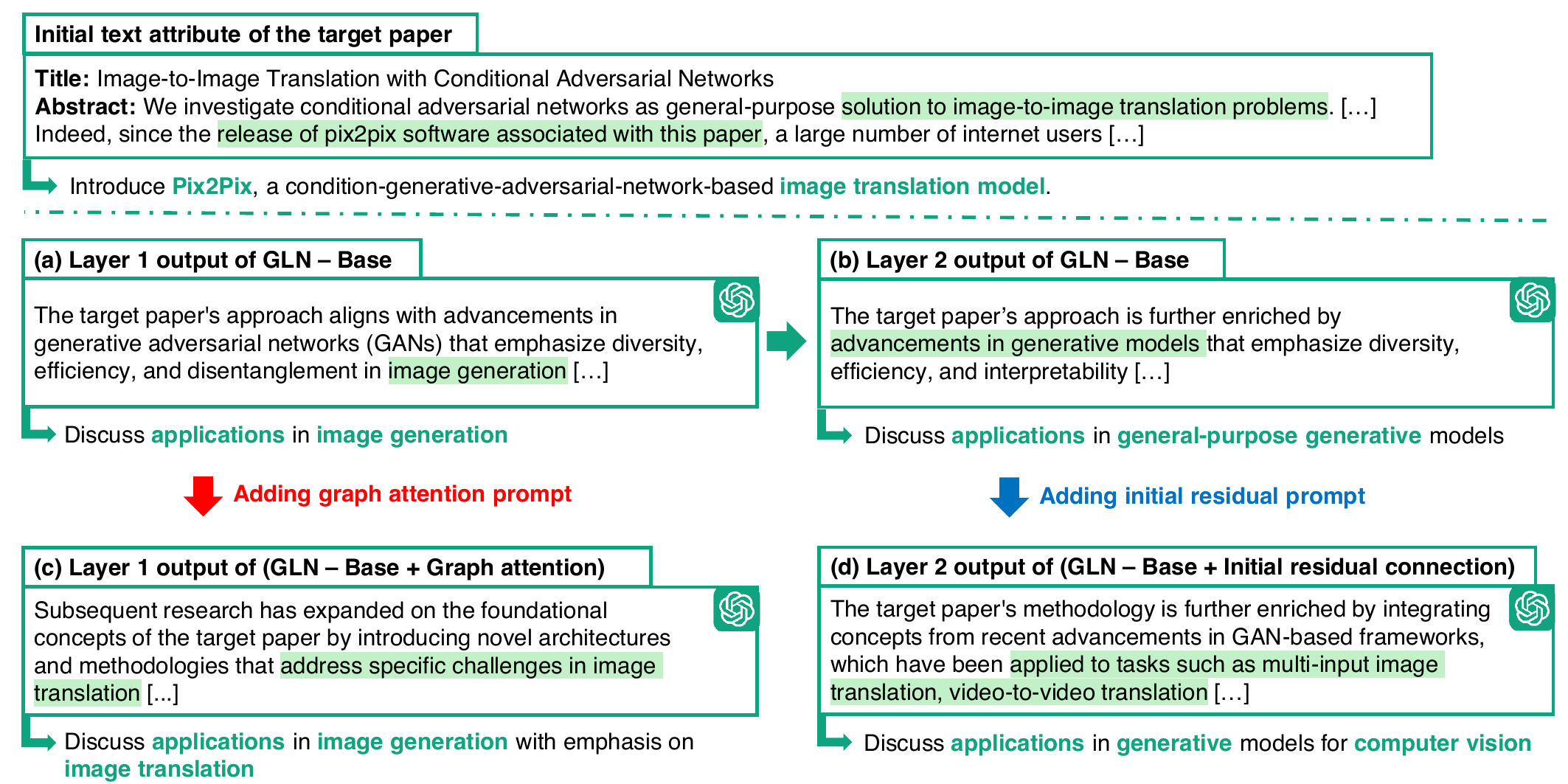}
    }
    \caption{A case study on~\citep{isola2017image}, showing \method-Base's representations of the paper at: (a) layer 1, (b) layer 2, (c) layer 1 with a graph attention prompt, and (d) layer 2 with an initial residual connection prompt.}  
    \label{fig:pix2pixcasestudy}  
\end{figure*}

\begin{figure*}[t!]
    \centering
    \fbox{
    \includegraphics[width=1.0\textwidth]{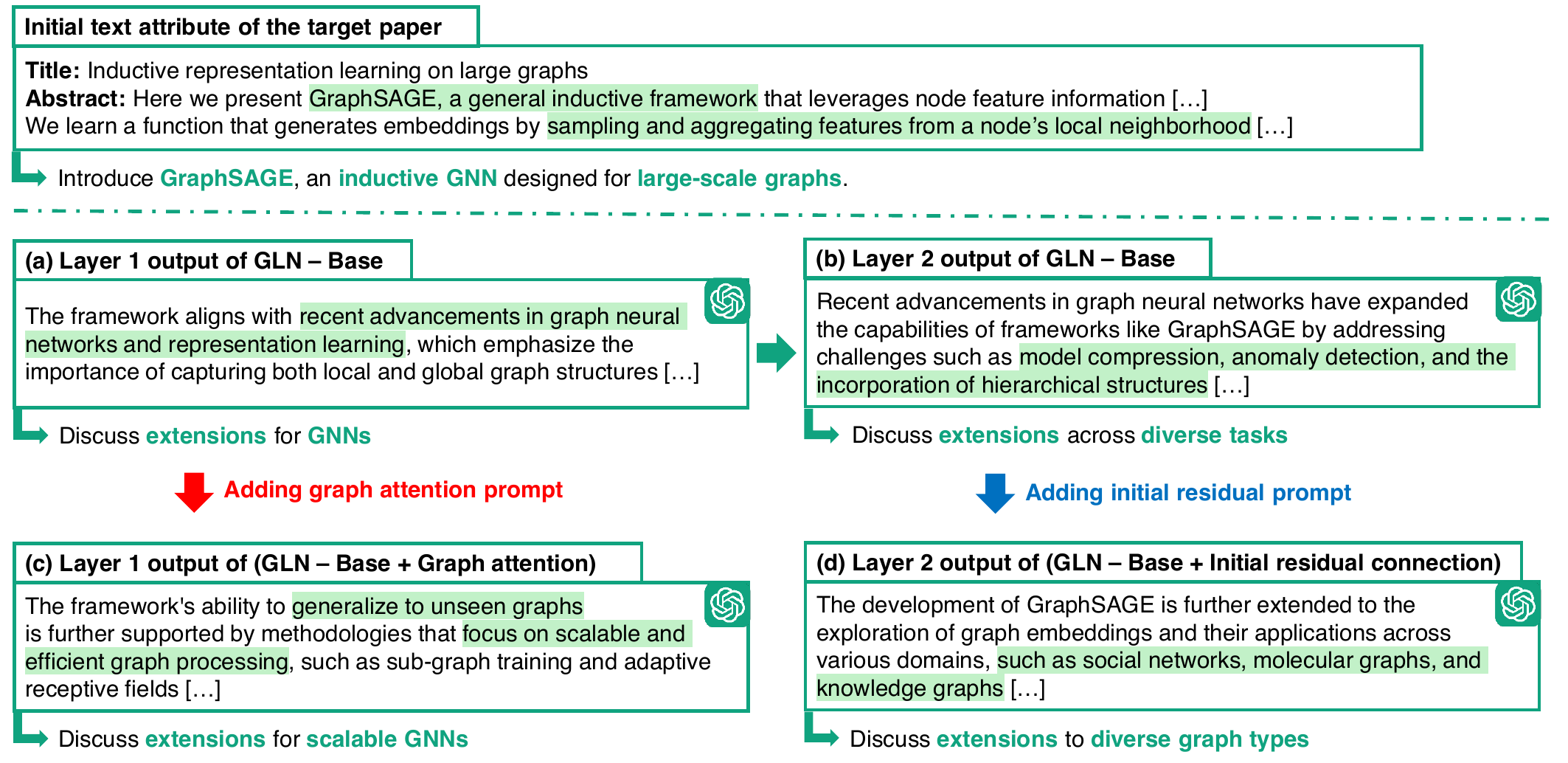}
    }
    \caption{A case study on~\citep{hamilton2017inductive}, showing \method-Base's representations of the paper at: (a) layer 1, (b) layer 2, (c) layer 1 with a graph attention prompt, and (d) layer 2 with an initial residual connection prompt.}  
    \label{fig:sagecasestudy}  
\end{figure*}

\begin{figure*}[t!]
    \centering
    \fbox{
    \includegraphics[width=1.0\textwidth]{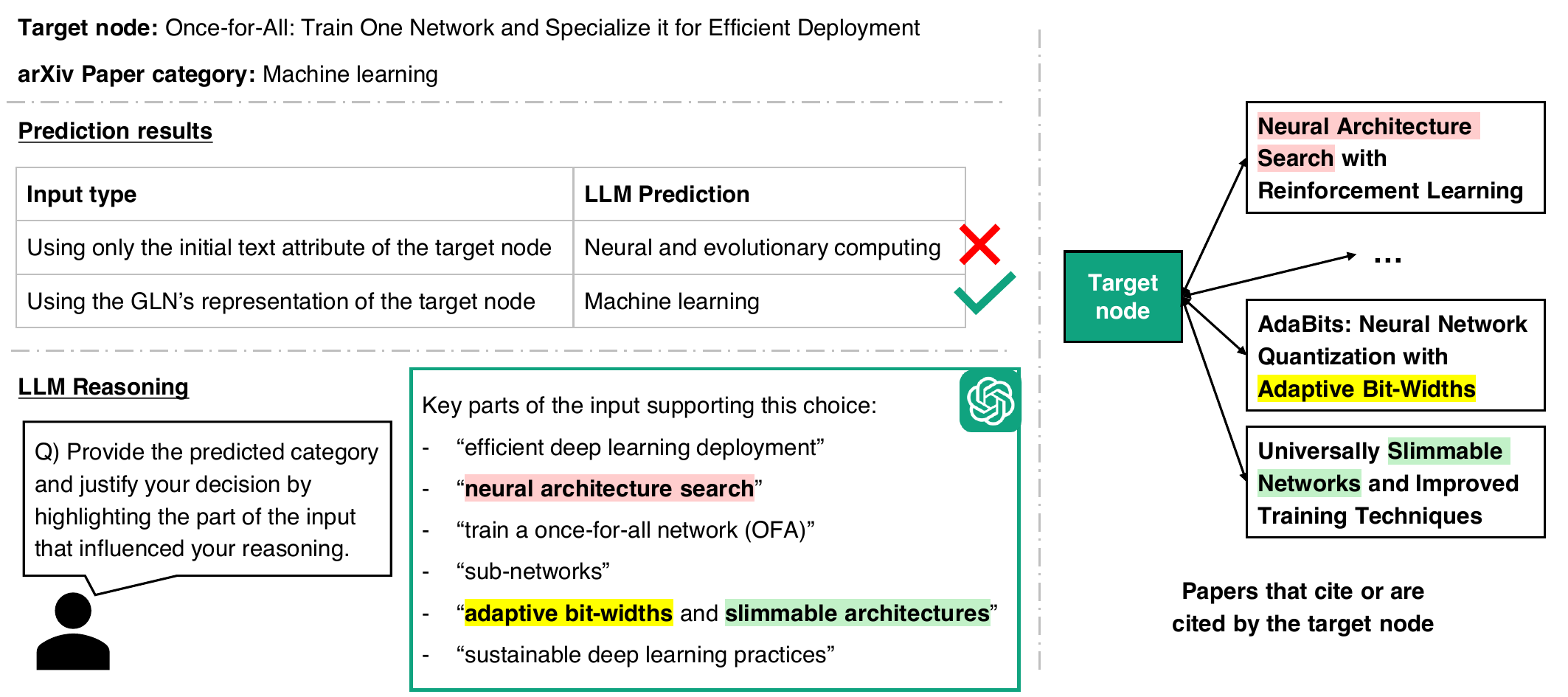}
    }
    \caption{A case study on~\citep{cai2019once}, showing LLM's reasoning for its downstream task decision.
    While an LLM misclassified the target node when only using the node attributes, it correctly classifies the target node by using information obtained from the target node's neighbors.}  
    \label{fig:reasoninganalysis}  
\end{figure*}

\begin{figure*}[t!]
    \centering
    \fbox{
    \includegraphics[width=1.0\textwidth]{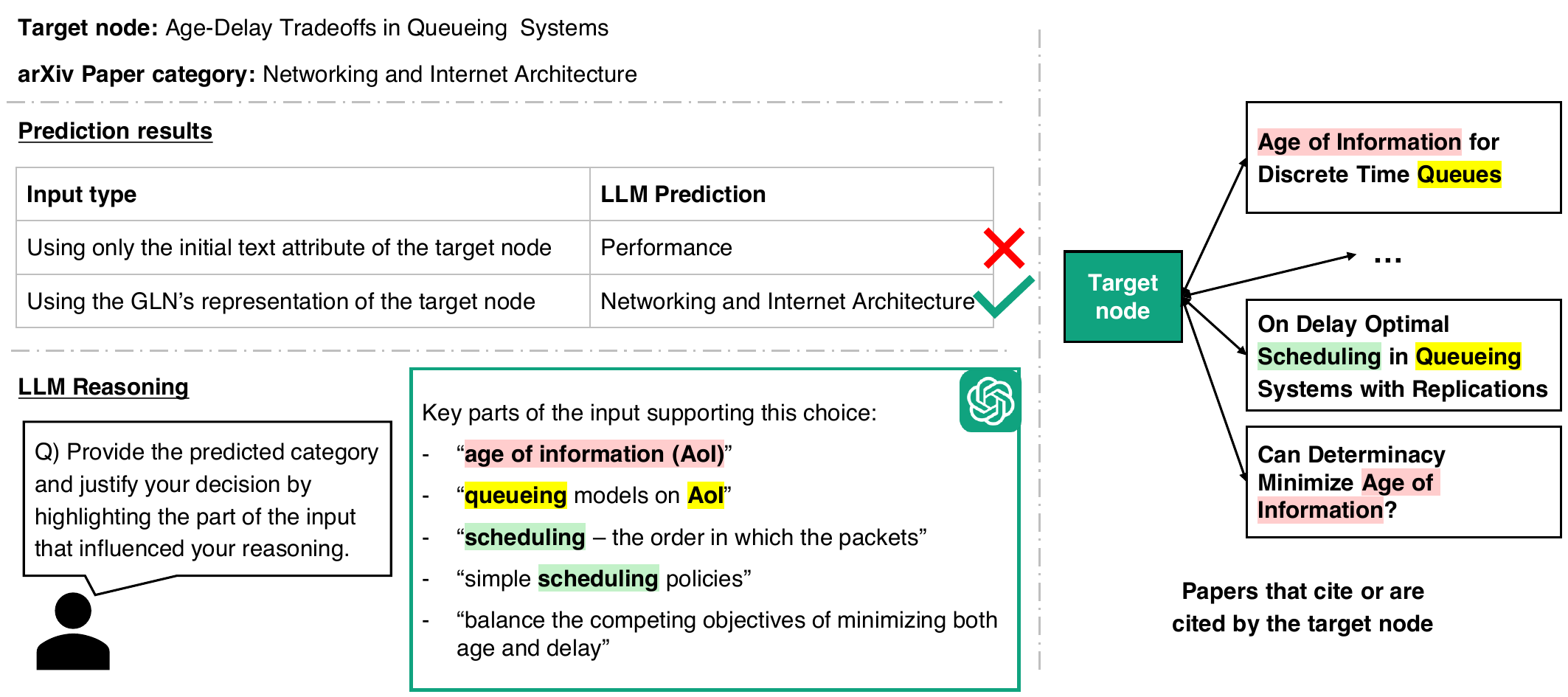}
    }
    \caption{A case study on~\citep{talak2020age}, showing LLM's reasoning for its downstream task decision.
    While an LLM misclassified the target node when only using the node attributes, it correctly classifies the target node by using information obtained from the target node's neighbors.}  
    \label{fig:reasoninganalysis2}  
\end{figure*}

In this appendix section, we provide additional experimental results that are omitted from the main section due to space constraints.
Specifically, we present two types of case studies:
Case studies analyzing representations of \method in various domains (Appendix~\ref{subapp:representationanalysis}) and case studies analyzing the reasoning of \method on downstream graph-related tasks (Appendix~\ref{subapp:llmreasoning}).

\subsection{Representation analysis}\label{subapp:representationanalysis}

We analyze the three popular papers from the three different domains:
\begin{itemize}[leftmargin=*]
    \item Natural language processing (NLP): ELMo, a pre-trained language model~\citep{peters2018deep}
    \item Computer vision (CV): Pix2Pix, an image translation generative model~\citep{isola2017image}
    \item Graph representation learning (GRL): GraphSAGE, an inductive graph neural network model~\citep{hamilton2017inductive}
\end{itemize}


\noindent\textbf{NLP Paper: ELMo.}
The case study result for ELMo~\citep{peters2018deep} is presented in Figure~\ref{fig:elmocasestudy}.
Below, we analyze whether the observations in Section~\ref{subsec:qualitative} are valid in \citep{peters2018deep}.
\begin{itemize}[leftmargin=*]
    \item \textit{\textbf{Observation 1.}} As shown in Figure~\ref{fig:elmocasestudy} (a) and (b), the layer-1 output focuses on the applications of context learning in {NLP}, while the layer-2 output extends to applications {beyond NLP}, such as scene graph generation and object detection.
    This result suggests that the representation gets more general across layers.
    \item \textit{\textbf{Observation 2.}} As shown in Figure~\ref{fig:elmocasestudy} (a) and (c), the output without the graph attention prompt lists applications of context learning in NLP, whereas the output with the prompt focuses on the integration of contextualized embeddings with additional features—a technique emphasized in the target paper.
    This result suggests that the representation gets specialized with the graph attention prompt.
    \item \textit{\textbf{Observation 3.}} As shown in Figure~\ref{fig:elmocasestudy} (b) and (d), the output without the initial residual connection prompt discusses the applications of context learning in various domains, while that with the initial residual connection prompt focuses on the architectural progress in NLP, domain where the target paper belongs to.
    This result suggests that the initial residual connection prompt helps maintain the information provided from the initial text attribute.
\end{itemize}
In summary, our analysis result suggests that the observations in Section~\ref{subsec:qualitative} are still valid in~\citep{peters2018deep}.


\noindent\textbf{CV Paper: Pix2Pix.}
The case study result for Pix2Pix~\citep{isola2017image} is presented in Figure~\ref{fig:pix2pixcasestudy}.
Below, we analyze whether the observations in Section~\ref{subsec:qualitative} are valid in \citep{isola2017image}.
\begin{itemize}[leftmargin=*]
    \item \textit{\textbf{Observation 1.}} As shown in Figure~\ref{fig:pix2pixcasestudy} (a) and (b), the layer-1 output focuses on the extensions of pix2pix in image generation, while the layer-2 output discusses its extensions for general generative models, without targeting specific domain.
    This result suggests that the representation gets more general across layers.
    \item \textit{\textbf{Observation 2.}} As shown in Figure~\ref{fig:pix2pixcasestudy} (a) and (c), the output without the graph attention prompt covers image generation, whereas the output with the prompt focuses on the image translation within image generation, a task the target paper focuses on.
    This result suggests that the representation gets specialized with the graph attention prompt.
    \item \textit{\textbf{Observation 3.}} As shown in Figure~\ref{fig:pix2pixcasestudy} (b) and (d), the output without the initial residual connection prompt discusses the general generative models, while that with the initial residual connection prompt focuses on the generative models for computer vision, which is the key domain the target paper belongs to.
    This result suggests that the initial residual connection prompt helps maintain the information provided from the initial text attribute.
\end{itemize}
In summary, our analysis result suggests that the observations in Section~\ref{subsec:qualitative} are still valid in~\citep{isola2017image}.


\noindent\textbf{GRL Paper: GraphSAGE.}
The case study result for GraphSAGE~\citep{hamilton2017inductive} is presented in Figure~\ref{fig:sagecasestudy}.
Below, we analyze whether the observations in Section~\ref{subsec:qualitative} are valid in \citep{hamilton2017inductive}.
\begin{itemize}[leftmargin=*]
    \item \textbf{\textit{Observation 1.}} 
    As shown in Figure~\ref{fig:sagecasestudy} (a) and (b), the layer-1 output focuses on the extensions of GraphSAGE for graph neural networks, while the layer-2 output covers the diverse applications of GraphSAGE, such as model compression and anomaly detection.
    This result suggests that the representation gets more general across layers.
    \item \textbf{\textit{Observation 2.}}
    As shown in Figure~\ref{fig:sagecasestudy} (a) and (c), the output without the graph attention prompt covers the extensions of GraphSAGE for general-purpose GNNs, while that with the prompt focuses on the inductive and/or scalable GNNs, which are key characteristics of GraphSAGE.
    This result suggests that the representation gets specialized with the graph attention prompt.
    \item \textit{\textbf{Observation 3.}} As shown in Figure~\ref{fig:sagecasestudy} (b) and (d), the output without the initial residual connection prompt discusses GraphSAGE applications across various tasks, whereas the output with the prompt emphasizes its use with different graph types, aligning with the target paper's broader focus on graph representation learning.
    This result suggests that the initial residual connection prompt helps maintain the information provided from the initial text attribute.
\end{itemize}
In summary, our analysis result suggests that the observations in Section~\ref{subsec:qualitative} are still valid in~\citep{hamilton2017inductive}.

\begin{figure*}[t!]
    \centering
    \fbox{
    \includegraphics[width=1.0\textwidth]{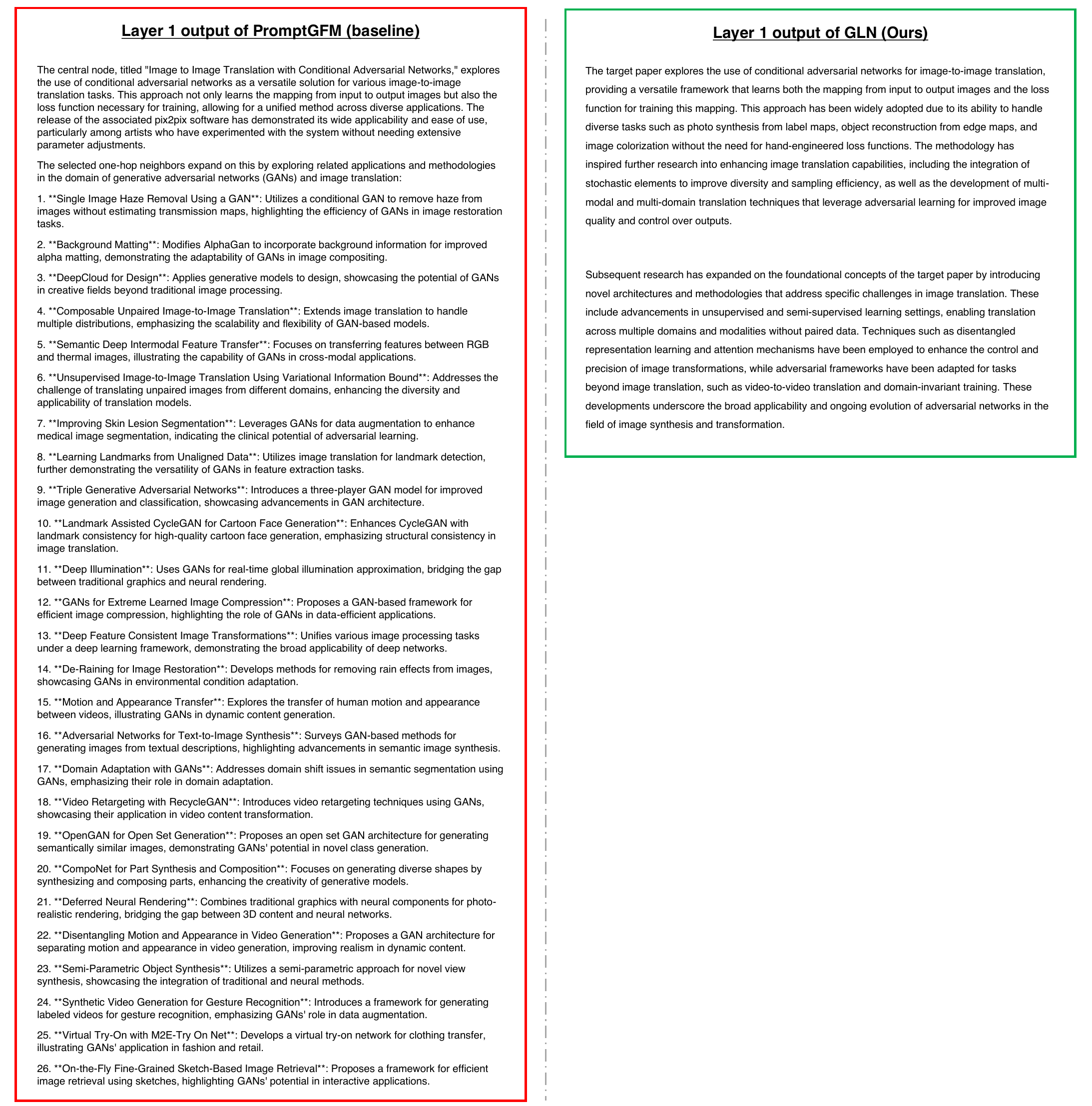}
    }
    \caption{A case study on~\citep{isola2017image}.
    While \texttt{PromptGFM} outputs a straightforward list of citing and cited papers, \method offers a comprehensible and succinct summary of those closely related to the target paper.}  
    \label{fig:promptgfmourspix2pix}  
\end{figure*}

\begin{figure*}[t!]
    \centering
    \fbox{
    \includegraphics[width=1.0\textwidth]{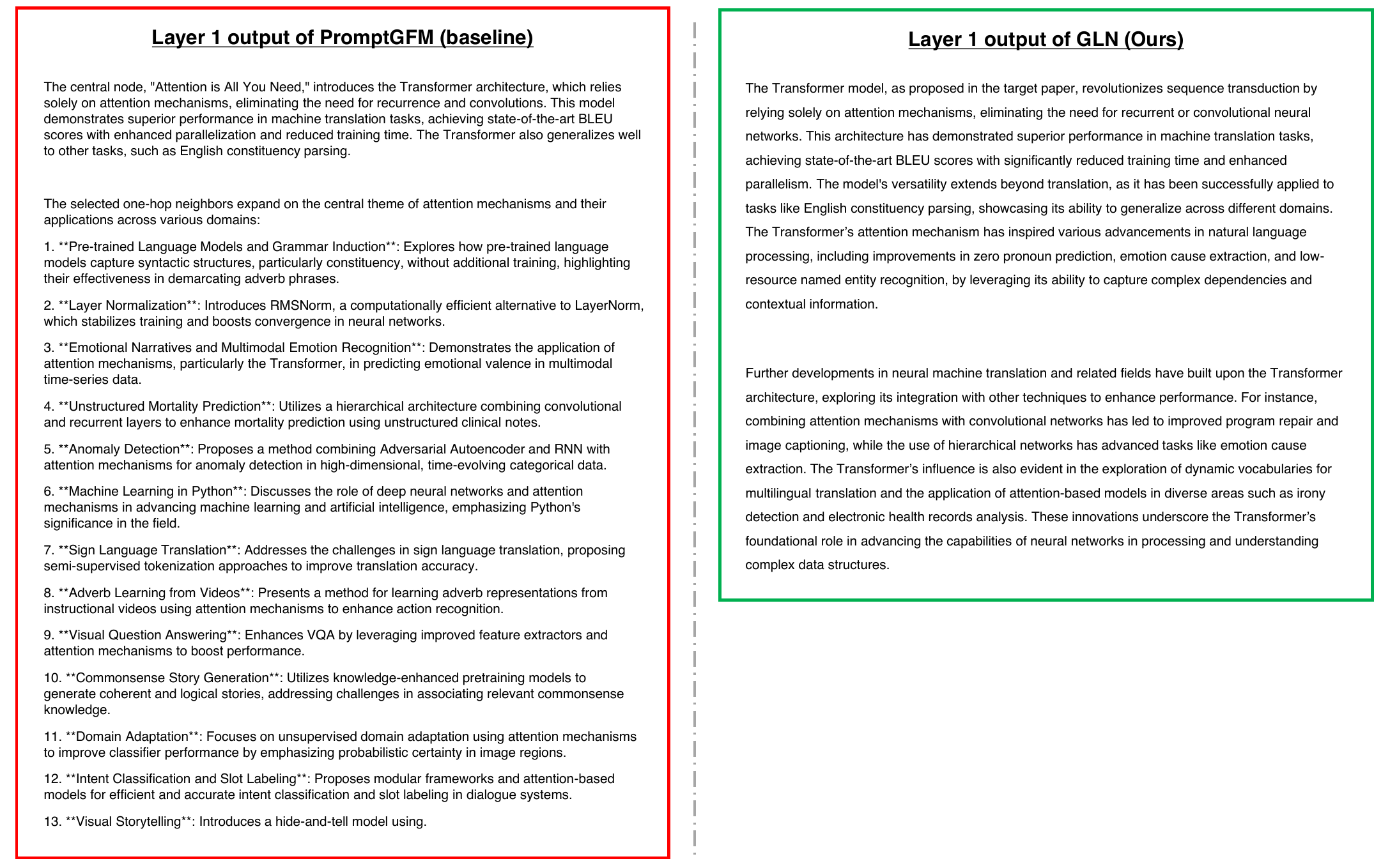}
    }
    \caption{A case study on~\citep{vaswani2017attention}.
    While \texttt{PromptGFM} outputs a straightforward list of citing and cited papers, \method offers a comprehensible and succinct summary of those closely related to the target paper.
    }  
    \label{fig:promptgfmtransformer}  
\end{figure*}

\begin{figure*}[t!]
    \centering
    \fbox{
    \includegraphics[width=1.0\textwidth]{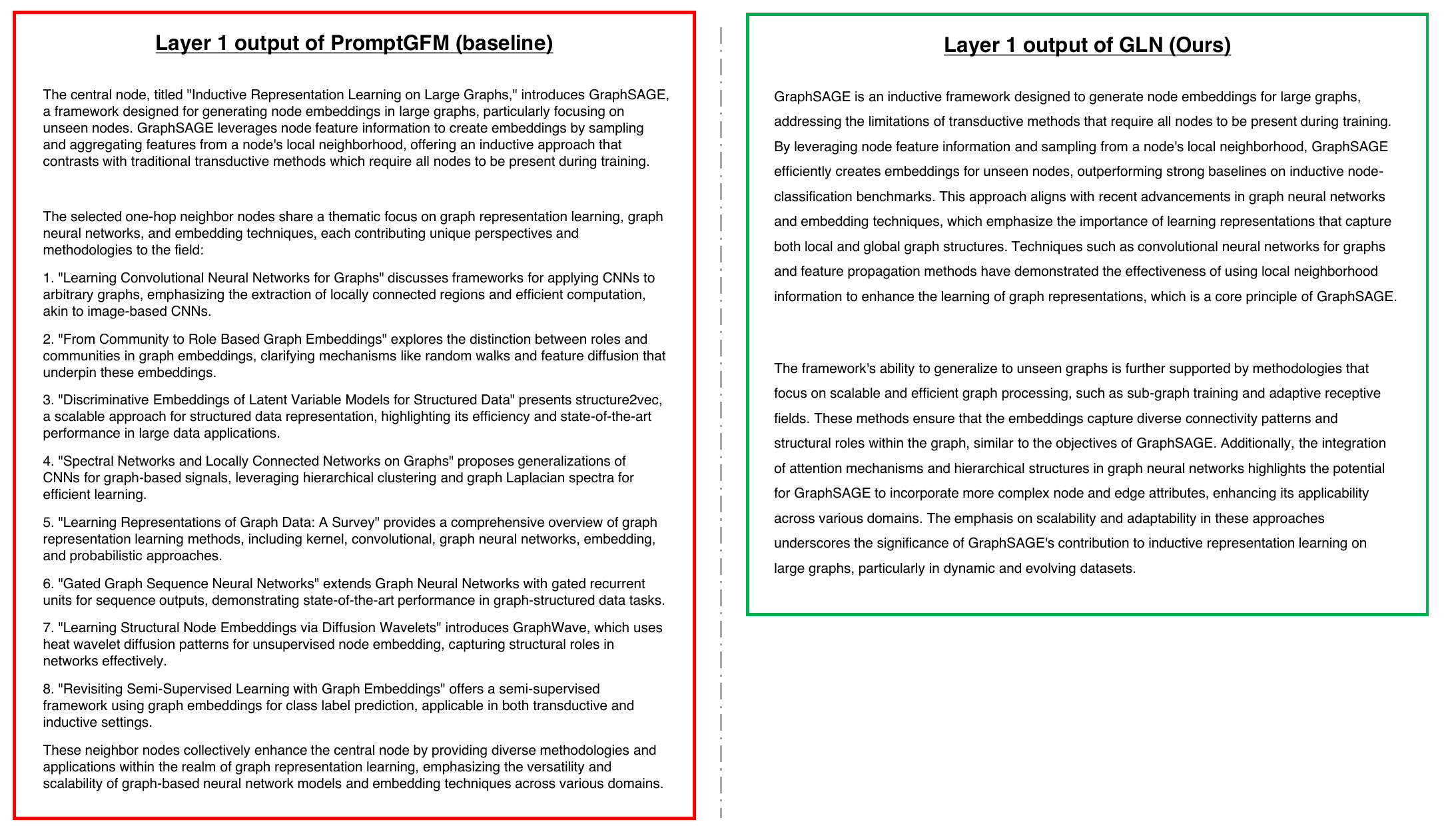}
    }
    \caption{A case study on~\citep{hamilton2017inductive}.
    While \texttt{PromptGFM} outputs a straightforward list of citing and cited papers, \method offers a comprehensible and succinct summary of those closely related to the target paper.}  
    \label{fig:promptgfmsage}  
\end{figure*}

\subsection{Reasoning analysis}\label{subapp:llmreasoning}

As noted in Section~\ref{sec:introduction}, textual representations allow an LLM to reason about its downstream task decisions.
In this section, we present case studies on two papers in the arXiv dataset where using only the initial text attribute results in misclassification, while using the representation from \method yields correct classification.

\noindent\textbf{Case study 1.}
The first case is about~\citep{cai2019once}.
As shown in Figure~\ref{fig:reasoninganalysis}, the LLM misclassifies the target node when relying solely on its initial textual attributes (title and abstract) but correctly classifies it when using the representation from \method, where the correct label is '\textit{machine learning}`. 
Notably, the LLM explicitly references terms from neighboring nodes (e.g., neural architecture search, slimmable networks) that are closely associated with the machine learning' category. 
This reasoning result suggests that integrating neighbor information can improve node classification performance, and \method effectively represents it.

\noindent\textbf{Case study 2.}
The second case is about~\citep{talak2020age}.
As shown in Figure~\ref{fig:reasoninganalysis}, the LLM misclassifies the target node when relying solely on its initial text attributes (title and abstract) but correctly classifies it when using the representation from \method, where the correct label is `\textit{networking and internet architecture}'. 
Notably, the LLM explicitly references terms from neighboring nodes (e.g., age of information, scheduling, and queueing) that are closely associated with the networking and internet architecture category. 
This reasoning result suggests that integrating neighbor information can improve node classification performance, and \method effectively represents it.

\subsection{Comparison with \texttt{PromptGFM}}\label{subapp:promptgfmours}
Recall that we briefly discussed the comparison between the outputs of \method and that of \texttt{PromptGFM}~\cite{zhu2025llm}, which is a baseline method (Section~\ref{subsec:relatedwork}).
In this section, we provide a detailed case study that compares (1) the representations obtained via \texttt{PromptGFM} and (2) those obtained via \method.

\noindent\textbf{Setup.} To this end, we present the first-layer outputs of \texttt{PromptGFM} and \method for three papers from different domains within the arXiv citation network: \citep{isola2017image} (CV), \citep{vaswani2017attention} (NLP), and \citep{hamilton2017inductive} (GRL).
We use GPT-4o as the backbone LLM for both methods.

\noindent\textbf{Results.}
In short, the output of \method is more comprehensible and well-structured, whereas that of \texttt{PromptGFM} is limited in its utility from a user comprehension perspective.
Specifically, as shown in Figure~\ref{fig:promptgfmourspix2pix}, representations from \texttt{PromptGFM} list brief summaries of papers that cite or are cited by the target paper~\citep{isola2017image}.
In contrast, \method returns a concise and focused summary of the target node’s neighbors, highlighting how generative adversarial networks (GANs) are applied to image translation tasks and further developed.

Similar results are shown in the outputs for \citep{vaswani2017attention} and \citep{hamilton2017inductive}, as shown in Figure~\ref{fig:promptgfmtransformer} and Figure~\ref{fig:promptgfmsage}, respectively.
These results suggest that \method yields clearer, better-structured outputs, whereas \texttt{PromptGFM}’s are far less useful for user comprehension.

\noindent\textbf{Potential reasons.}
We hypothesize that the differences in user comprehensibility primarily arise from the specific \textbf{\textit{task}} assigned to the LLM.
Specifically, \texttt{PromptGFM} prompts an LLM to `\textit{\textbf{aggregate} neighbor nodes and \textbf{update}
a concise yet meaningful representation for the central node}'.
This prompt likely leads the LLM to focus heavily on aggregating neighbor information, resulting in a mere enumeration of the target node’s neighbors.

In contrast, in \method, we prompt an LLM to \textit{refine} the target node's description \textit{by incorporating} its neighbor information.
This guides the LLM to center its attention on the target node and produce a target-node-centric summary of its neighbors, improving the user comprehensibility of the output.

\subsection{Ablation study}\label{subapp:ablation}
In this section, we provide further ablation studies of \method: demonstrating whether each (1) graph attention prompt and (2) initial residual connection prompt is effective for the downstream task.

As shown in Table~\ref{tab:ablationstudy}, \method—which incorporates both graph attention and initial residual connection prompts—outperforms all three variants that omit either or both prompts.
This result suggests that the two advanced GNN-style prompts are essential for good downstream task performance.

\begin{table}[t]
    \centering
    \small
    \begin{tabular}{cc|cc|cc}
        \toprule
         \multirow{2}{*}{G.A.} & \multirow{2}{*}{I.R.C.} & \multicolumn{2}{c|}{OGBN-arXiv} & \multicolumn{2}{c}{Book-History} \\

         &  & Node. & Link. & Node. & Link. \\
        \midrule
        \midrule
        \xmark & \xmark & 63.0 & 92.4 & 45.8 & 86.4 \\
        
        \cmark & \xmark & 62.8 & 92.4 & 47.0 & 86.6 \\
        
        \xmark & \cmark & 63.5 & 92.2 & 46.7 & 86.4 \\
        
        \cmark & \cmark & \textbf{64.0} & \textbf{93.0} & \textbf{47.3} & \textbf{87.0} \\
        \bottomrule
    \end{tabular}
    \caption{\textbf{Graph attention prompt and initial residual connection prompt are essential for strong performance.} Ablation study result of \method.
    G.A. and I.R.C. denote the graph attention prompt and the initial residual prompt connection prompt, respectively.
    In addition, Node. and Link. denote node classification and link prediction, respectively.
    The best performance in each setting is highlighted in \textbf{bold.}}
    \label{tab:ablationstudy}
\end{table}

\begin{table}[t]
    \centering
    \small
    \begin{tabular}{l|c|c}
        \toprule
         Models & {OGBN-arXiv} & {Wiki-CS} \\

        \midrule
        \midrule
        
        \method orig. & 77.5 &  82.5 \\
        
        \method w/o GA & 75.5 & 79.0 \\
        
        \method w/ GA & 76.5 & 81.5 \\

        \bottomrule
    \end{tabular}
    \caption{\textbf{Graph attention prompt helps \method to maintain its performance under edge corruption.} The node classification performance comparison under input noise.
    \method w/o GA and w/ GA indicate the \method performance under edge corruption without and with the graph attention prompt, respectively.
    \method orig. indicates the performance of \method on the original datasets.
    }
    \label{tab:graphattention}
\end{table}

\subsection{Analysis of the graph attention prompt}\label{subapp:graphattention}

In this section, we provide an in-depth analysis regarding the effect of the graph attention prompt, which is used in \method. 
Recall that our key intuition behind the graph attention prompt is to instruct the LLM to focus on the relevant neighbors of the target node. 
To validate this, we corrupt each node’s neighborhood by injecting random neighbors and examine whether the graph attention prompt helps \method maintain performance under such neighbor corruption.
Specifically, for each node, we sample 7 of its ground-truth neighbors and add 3 nodes sampled from the entire graph, yielding a 10-neighbor set fed to \method.
For experiments, we sample 200 target nodes from the OGBN-ArXiv and Wiki-CS datasets and use a 1-layer \method with GPT-4.1-nano as the backbone encoder and GPT-4.1 as the downstream-task-performing LLM.

As shown in Table~\ref{tab:graphattention}, \method without the graph attention prompt suffers a significant performance drop, whereas \method with the prompt maintains performance to some extent. This suggests that the graph attention prompt helps filter the relevant neighbors of the target node.

\subsection{Improving scalability of \method}\label{subapp:scalability}

In this section, we analyze several strategies that can improve the scalability of \method. 
Specifically, we explore three approaches: (1) use of a small language model (SLM), (2) use of an input-token efficient strategy, and (3) use of an output-token efficient strategy.
To this end, we sample 200 nodes from the OGBN-ArXiv and Wiki-CS datasets and use a 1-layer \method with GPT-4.1-nano as the backbone encoder and GPT-4.1 as the downstream-task-performing LLM.

\begin{table}[t]
    \centering
    \small
    \begin{tabular}{l|c|c}
        \toprule
         Models & {OGBN-arXiv} & {Wiki-CS} \\

        \midrule
        \midrule
        
        PromptGFM with LLM & 75.0 (6.4) & 79.5 (7.6) \\
        
        \method with LLM & 78.0 (8.2) & 82.5 (7.1) \\
        
        \method with SLM & 77.5 (2.4) & 82.5 (2.2) \\
        
        \bottomrule
    \end{tabular}
    \caption{\textbf{\method achieves high speed and effectiveness with an SLM.} The node classification performance comparison under diverse-sized LLMs.
    Numbers in parentheses indicate the average encoding time per node.}
    \label{tab:variousllms}
\end{table}

\noindent\textbf{Effectiveness under SLM.} 
In \method, replacing the LLM with an SLM can improve scalability, since SLMs require significantly less generation time.
To evaluate model performance, we compare \method equipped with SLM (GPT-4.1-nano) against (1) \method equipped with an LLM (GPT-4.1) and (2) PromptGFM equipped with an LLM (GPT-4.1).

As shown in Table~\ref{tab:variousllms}, \method with SLM requires less than one-third the encoding time of \method with LLM, while maintaining comparable performance.
In addition, \method outperforms PromptGFM even with a smaller backbone language model.
This result demonstrates that the scalability of \method can be improved by using SLM, while being effective.

\begin{table}[t]
    \centering
    \small
    \begin{tabular}{l|c|c}
        \toprule
         Models (\# of neighbors) & {OGBN-arXiv} & {Wiki-CS} \\

        \midrule
        \midrule
        
        PromptGFM ($N=10$) & 73.5 (2772) & 78.5 (9298) \\
        
        \method ($N=3$) & 76.0 (1067) & 81.5 (3419) \\
        
        \method ($N=5$) & 76.5 (1423) & 81.5 (4649) \\

        \method ($N=10$) & 77.5 (2690) & 82.5 (8975) \\
        
        \bottomrule
    \end{tabular}
    \caption{\textbf{\method remains strong under fewer neighbor samples.} The node classification performance comparison under diverse-sized neighbor samples.
    Numbers in parentheses indicate the average number of input tokens of each case.}
    \label{tab:neighborsample}
\end{table}

\noindent\textbf{Input-prompt efficient strategy}
In \method, we sample a fixed number of neighbors instead of utilizing all available ones, as detailed in Appendix~\ref{subapp:settingdetail}.
We investigate how varying the number of neighbor samples influences \method
encoding.

As shown in Table~\ref{tab:neighborsample}, \method outperforms PromptGFM even when using significantly fewer neighbor samples—and thus, far fewer input tokens. 
This result demonstrates that strong performance can be achieved with substantially fewer input tokens than required by the baseline method.

\begin{table}[t]
    \centering
    \small
    \begin{tabular}{l|c|c}
        \toprule
         Models (output constraint) & {OGBN-arXiv} & {Wiki-CS} \\

        \midrule
        \midrule
        
        PromptGFM (N/A) & 73.5 (393) & 78.5 (491) \\
        
        \method (2-paragraphs) & 77.5 (256) & 82.5 (257) \\
        
        \method (3-sentences) & 77.0 (110) & 81.5 (109) \\

        \bottomrule
    \end{tabular}
    \caption{\textbf{\method remains strong under stricter output-length constraint.} The node classification performance comparison under several output constraints.
    Numbers in parentheses indicate the average number of output tokens of each case.}
    \label{tab:outputconstraint}
\end{table}

\noindent\textbf{Output-prompt efficient strategy}
We limit the output representation of \method by 2 paragraphs, as detailed in Appendix~\ref{subapp:glnencoding}.
To further reduce the output length, we prompt the LLM to generate a shorter response, constrained to 3 sentences.

As shown in Table~\ref{tab:outputconstraint}, \method outperforms the baseline method even when the output length is constrained at the prompt level. 
This result demonstrates that strong performance can be achieved with substantially fewer output tokens than those used by the baseline methods.

\begin{table}[t]
    \centering
    \small
    \begin{tabular}{l|c|c}
        \toprule
         Models & {OGBN-arXiv} & {Wiki-CS} \\

        \midrule
        \midrule
        
        \method orig. & 77.5 &  82.5 \\
        
        \method w/o denoising & 75.5 &  81.0 \\
        
        \method w/ denoising & 77.0 & 82.5 \\

        \bottomrule
    \end{tabular}
    \caption{\textbf{Performance of \method decreases under input noises, while a simple text denoising technique can mitigate this.} The node classification performance comparison under input noise.
    \method w/o denoising and w/ denoising indicate the performance of \method on noisy datasets without and with the application of denoising techniques, respectively.
    \method orig. indicates the performance of \method on the original datasets.
    }
    \label{tab:noisy}
\end{table}

\subsection{Analysis under noisy node attributes}\label{subsec:noisyinput}

In this section, we investigate the effectiveness of \method when the input node text attributes contain noise.
To this end, we randomly remove 30\% of the words from each node text attribute.
In addition, to examine whether denoising improves performance under noise, we apply a simple denoising technique: (1) an LLM is prompted to denoise the input node attribute by extracting the key concept of the given text, and (2) the resulting denoised text is then used as the node attribute.
For experiments, we sample 200 target nodes from the OGBN-ArXiv and Wiki-CS datasets and use a 1-layer \method with GPT-4.1-nano as the backbone encoder and GPT-4.1 as the downstream-task-performing LLM.

As shown in Table~\ref{tab:noisy}, (1) the performance of \method decreases when input node attributes are noisy, while (2) applying a denoising technique alleviates this performance degradation. 
These results suggest that although \method is sensitive to noise in node text attributes, adequate denoising can effectively mitigate such a negative impact.

\section{Experiment details}
\label{appendix:experimentdetail}
In this appendix section, we provide experimental details omitted from the main paper (Section~\ref{sec:analysis}).

\subsection{Experimental setting details}\label{subapp:settingdetail}

We describe the detailed experimental setting of the two downstream tasks: (1) node classification and (2) link prediction.

\noindent\textbf{Node classification.}
For node classification, we sample 1,000 nodes with degrees greater than or equal to 10 from each dataset (i.e., $\{v_{i} \in \mathcal{V} : \vert \mathcal{N}_{i}\vert \geq 10\}$).
We then obtain the textual representations of the corresponding nodes and prompt an LLM to predict their classes.
Lastly, measure accuracy by comparing the predicted classes with the ground-truth classes.

\noindent\textbf{Link prediction.}
For link prediction, we sample 500 edges whose endpoint nodes each have a degree greater than 10 (i.e., $\{e_{i} = \{v_{s},v_{t}\} \in \mathcal{E} : \vert \mathcal{N}_{s} \vert , \vert \mathcal{N}_{t} \vert \geq 10\}$).
We then remove these edges from $\mathcal{E}$ and obtain the textual representations of their endpoint nodes.
Next, for each edge, we: (1) randomly sample four nodes not connected by the edge using rule-based sampling, (2) provide one endpoint of the edge as input to the LLM, (3) construct a candidate set consisting of the true other endpoint and the four sampled nodes, and (4) prompt the LLM to select the node most likely to be linked with the given node.
Lastly, we measure the Hit-ratio@1 for edges, which is defined as $\frac{1}{\vert \mathcal{E}'\vert }\sum_{e_{i} \in \mathcal{E}'} \mathbf{1}[\texttt{LLM}(e_{i})]$, where $\mathcal{E}'$ is a set of sampled edges and $\mathbf{1}[\texttt{LLM}(e_{i})]$ is an indicator function that returns 1 if the LLM correctly predicts the another endpoint of $e_{i}$, otherwise 0.

\subsection{Baseline and \method details}\label{subapp:baselinedetail}

We describe the detailed setting of each method, including baseline methods and \method.

\noindent\textbf{{LLMs for downstream tasks.}} 
We found that in downstream tasks, GPT-4o-mini and Claude-3.0-Haiku—used as our backbone LLMs—often fail to return outputs in the assigned format, making automatic evaluation challenging.
Therefore, only for downstream tasks, we used more up-to-date models.
Specifically, we used GPT-4.1-mini and Claude-3.5-Haiku instead of GPT-4o-mini and Claude-3.0-Haiku, respectively.

\noindent\textbf{\texttt{Direct}.}  
This method performs the downstream task using only the target node's initial text attribute, without modifying its textual representation through certain LLM operations.

\noindent\textbf{\texttt{All-in-One}.} 
This is our newly introduced baseline that directly prompts an LLM to refine the target node's representation using its (1) one-hop and (2) two-hop neighbors. 
Due to input length constraints of LLMs, we sample 10 one-hop neighbors and 20 two-hop neighbors, uniformly at random, and provide them as input to the LLM.

\noindent\textbf{\texttt{PromptGFM}, \method-Base, and \method.}
For these methods, which leverage message passing, we stack 2 layers, which is a conventional setting in GNN research~\cite{kipf2017semi, velivckovic2017graph, hamilton2017inductive}.
Due to input length constraints of LLMs, we sample 10 neighbors for each node, uniformly at random, and use them for message passing.

\section{Prompt details}
\label{appendix:prompt}

In this appendix section, we provide detailed prompts used for (1) the encoding process of \method and (2) zero-shot downstream tasks (i.e., node classification and link prediction).

\subsection{Prompt for \method's encoding}\label{subapp:glnencoding}


\begin{figure*}[t!]
    \centering
    \fbox{
    \includegraphics[width=1.0\textwidth]{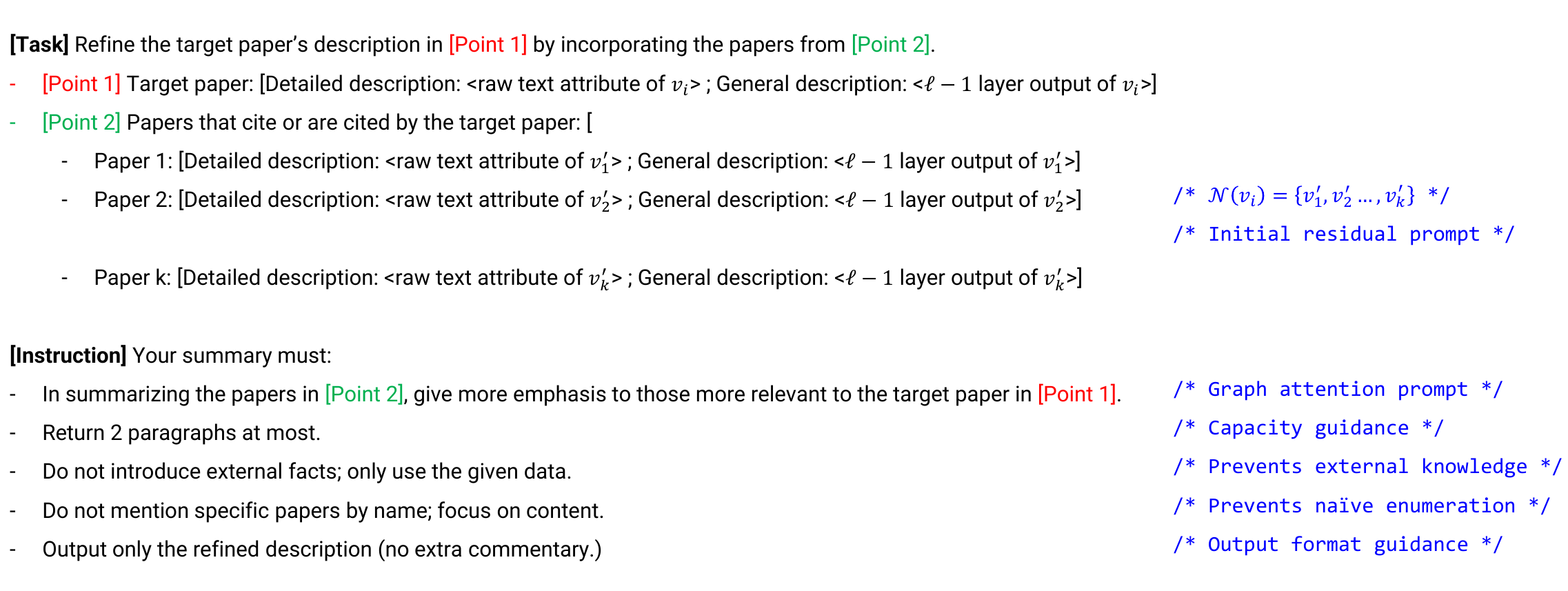}
    }
    \caption{Example prompt of \method for citation networks (oGBN-arXiv dataset).}  
    \label{fig:arxivprompt}  
\end{figure*}

\begin{figure*}[t!]
    \centering
    \fbox{
    \includegraphics[width=1.0\textwidth]{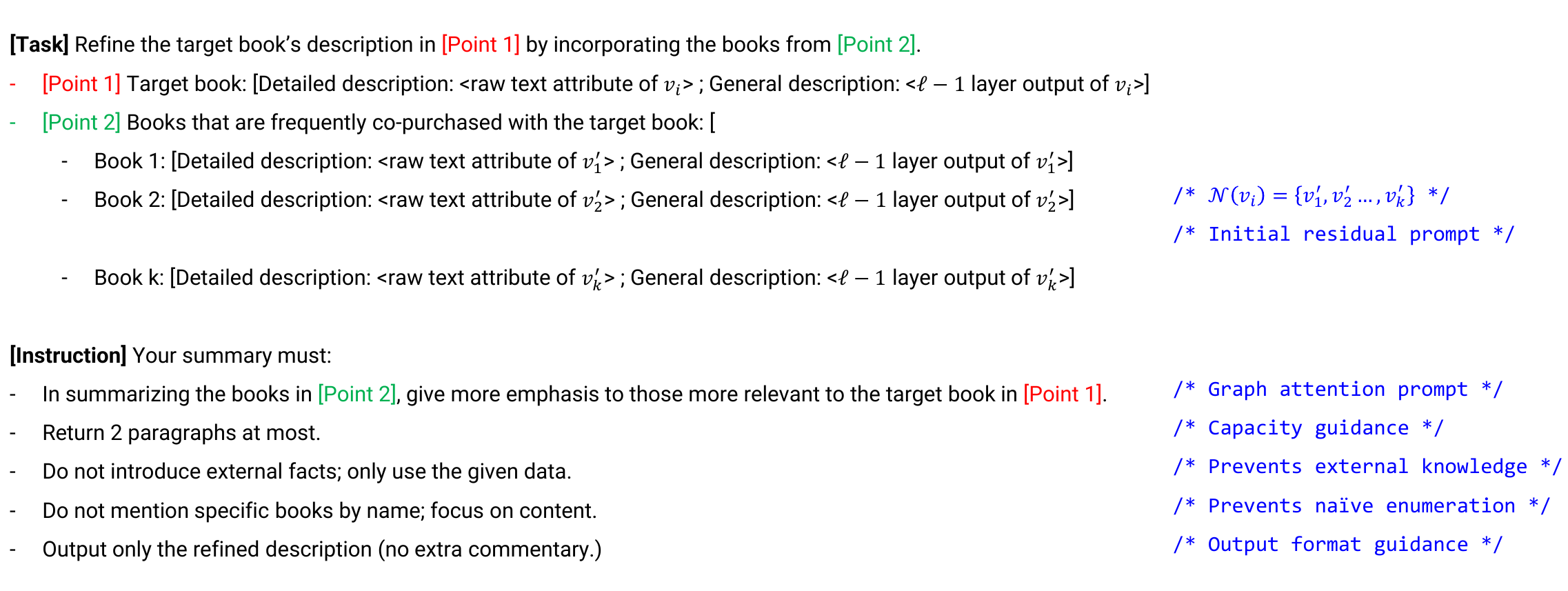}
    }
    \caption{Example prompt of \method for co-purchase networks (Book-History dataset).}  
    \label{fig:bookprompt}  
\end{figure*}

\begin{figure*}[t!]
    \centering
    \fbox{
    \includegraphics[width=1.0\textwidth]{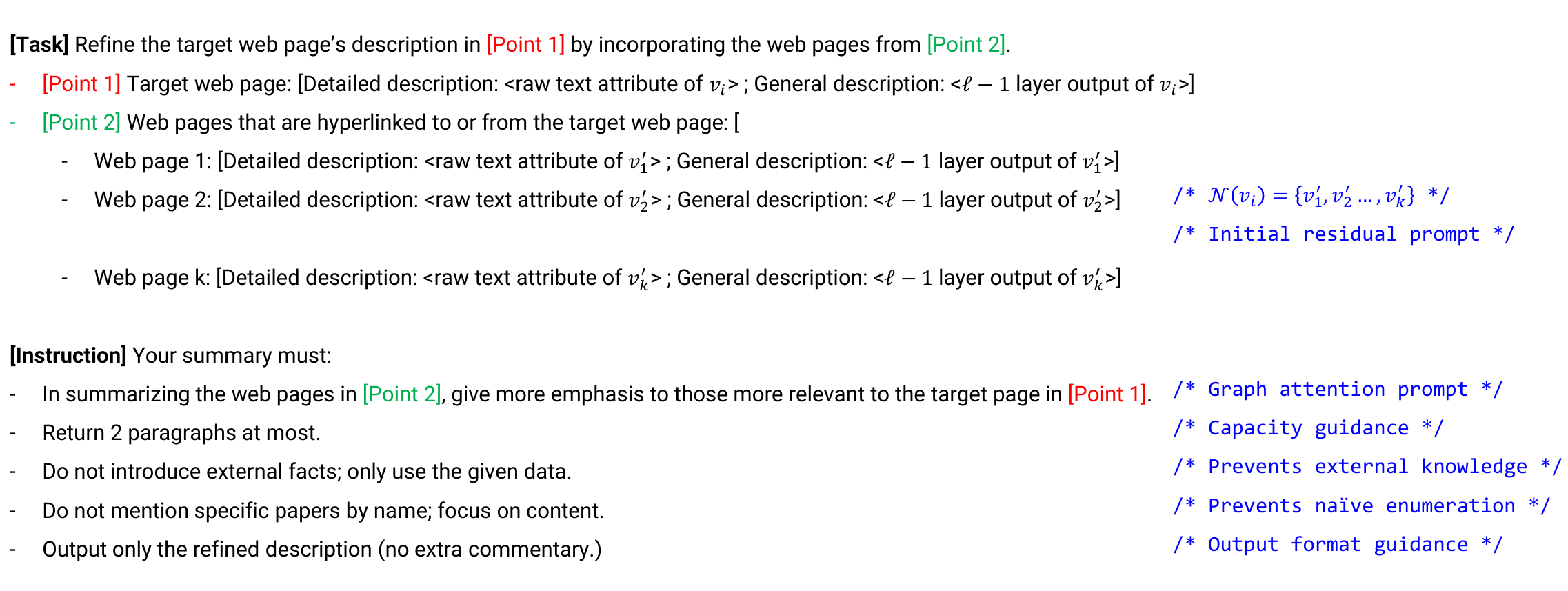}
    }
    \caption{Example prompt of \method for hyperlink networks (Wiki-CS dataset).}  
    \label{fig:pageprompt}  
\end{figure*}

\noindent\textbf{Prompt design.} 
We provide a detailed prompt design of \method. 
Specifically, we present the following types of prompts:
\begin{itemize}[leftmargin=*]
    \item \textbf{Prompt for citation networks}: A prompt for citation networks is in Figure~\ref{fig:arxivprompt}.
    \item \textbf{Prompt for co-purchase networks}: A prompt for book co-purchase networks is in Figure~\ref{fig:bookprompt}.
    \item \textbf{Prompt hyperlink networks}: A prompt for hyperlink networks is in Figure~\ref{fig:pageprompt}.
\end{itemize}

\begin{table}[t]
    \centering
    \small
    \begin{tabular}{l|c|c}
        \toprule
         Models & {OGBN-arXiv} & {Wiki-CS} \\

        \midrule
        \midrule
        
        \method orig. & 78.0 &  83.0 \\
        
        \method w/ new GA & 78.0 & 82.5 \\
        
        \method w/ new IRC & 76.0 & 80.5 \\
        
        \bottomrule
    \end{tabular}
    \caption{\textbf{Graph attention prompt is less sensitive to the choice of the attention-related phrase, while itemization for initial residual connection is necessary for the performance.} The node classification performance comparison under input noise.
    \method w/ new GA and w/ new IRC indicate \method equipped with a new graph attention prompt and a new initial residual connection prompt, respectively.
    \method orig. indicates the performance of \method with its original prompt design.
    }
    \label{tab:promptalternative}
    \vspace{-2mm}
\end{table}

\noindent\textbf{Investigating alternatives.}
We further analyze the prompt-robustness of \method. 
Specifically, we analyze our prompt designs for (1) the graph attention prompt and (2) the initial residual connection prompt. 
The core of the graph attention prompt lies in the \textit{phrase: `give more emphasis to those more relevant to the target'}.
Analogously, the core of the initial residual connection prompt lies in its \textit{itemized structure}, which explicitly distinguishes between the input node attributes and the outputs of the preceding layers.

To validate the effectiveness of such designs, we use a new graph attention prompt that uses \textit{`weigh highly the works most closely related to the target'} instead of the phrase mentioned above.
In addition, we also use an alternative initial residual connection prompt that uses the plain-text prompt instead of an itemization-based prompt.
Specifically, instead of using the itemized structure described in Figure~\ref{fig:arxivprompt}, we use: \textit{`The detailed description is [...]. The version updated by papers that cite or are cited by it is [...]'} for the initial residual connection.

As shown in Table~\ref{tab:promptalternative}, performance under an alternative graph attention prompt remains largely unchanged, but declines markedly when the itemized initial residual connection prompt is substituted with a plain-text description.
This indicates that the graph attention prompt is relatively insensitive to the exact phrasing as long as the attention objective is preserved, whereas explicitly itemized structure is indispensable for the initial residual connection prompt for high performance.

\subsection{Representation format of \method}\label{subapp:representationformat}
In this section, we further elaborate on the detailed format of the target node's representation produced by \method.
Specifically, we present a format for a citation network.

\noindent\colorbox{lightgreen}{%
\small
  \parbox{\linewidth}{%
    \texttt{Paper: \{\\
    - Detailed description: <initial text attribute>, \\
    - General description: <layer-1 output of \method>, \\
    - Highly general description: <layer-2 output of \method>\}
    }
  }%
}
This format is provided as a representation for the target node (paper). 
In the co-purchase dataset (Book-History) and hyperlink dataset (Wiki-CS), we use the terms `\texttt{Book}' and `\texttt{Web page}', instead of \texttt{Paper}, respectively.


\begin{figure*}[t!]
    \centering
    \fbox{
    \includegraphics[width=1.0\textwidth]{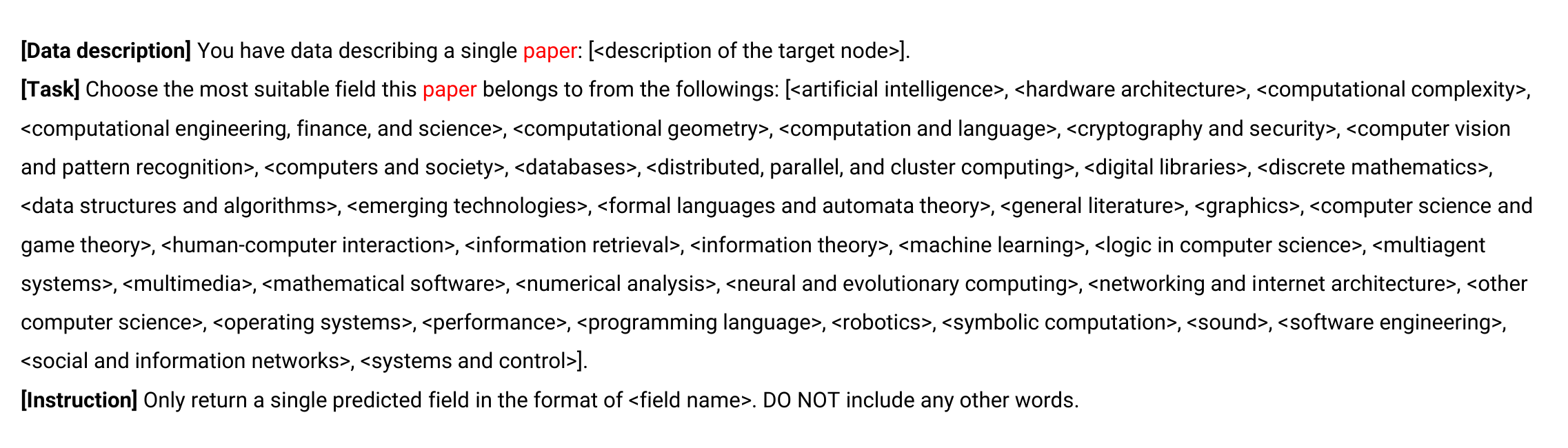}
    }
    \caption{Example node classification prompt of \method for citation networks (OGBN-arXiv dataset).}  %
    \label{fig:arxivnodeprompt}  
\end{figure*}

\begin{figure*}[t!]
    \centering
    \fbox{
    \includegraphics[width=1.0\textwidth]{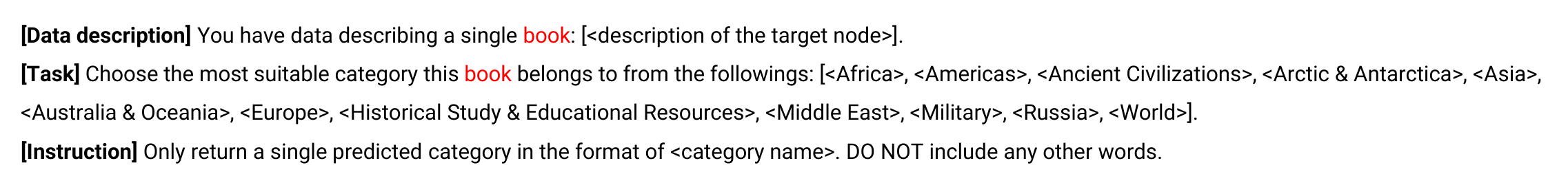}
    }
    \caption{Example node classification prompt of \method for co-purchase networks (Book-History dataset).}  
    \label{fig:booknodeprompt}  
\end{figure*}

\begin{figure*}[t!]
    \centering
    \fbox{
    \includegraphics[width=1.0\textwidth]{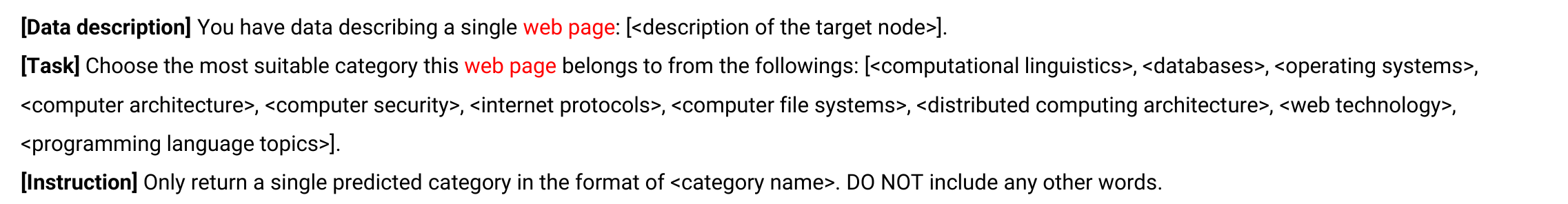}
    }
    \caption{Example node classification prompt of \method for hyperlink networks (Wiki-CS dataset).}  
    \label{fig:pagenodeprompt}  
\end{figure*}

\subsection{Prompt for downstream tasks}\label{subapp:detailtaskprompt}

In this section, we provide details regarding our prompt for downstream tasks, which are node classification and link prediction.

\noindent\textbf{Node classification}
Example node classification prompts for the OGBN-arXiv dataset (citation network), Book-History (co-purchase), and Wiki-CS (hyperlink network), are provided in Figure~\ref{fig:arxivnodeprompt}, \ref{fig:booknodeprompt}, and \ref{fig:pagenodeprompt}, respectively.
Specifically, we provide a set of possible categories and ask the LLM to select the one the target node is most likely to belong to.

\begin{figure*}[t!]
    \centering
    \fbox{
    \includegraphics[width=1.0\textwidth]{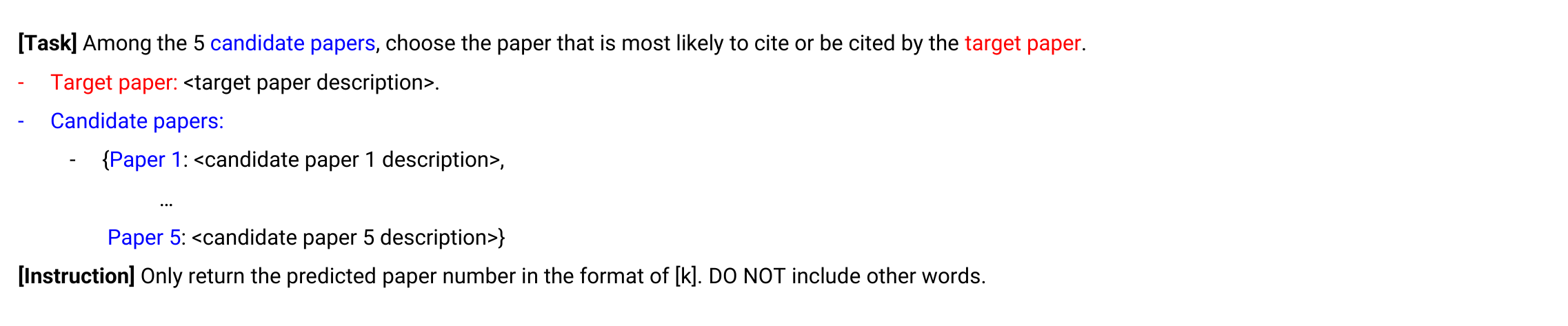}
    }
    \caption{Example edge prediction prompt of \method for citation networks (OGBN-arXiv dataset).}  
    \label{fig:arxivedgeprompt}  
\end{figure*}

\begin{figure*}[t!]
    \centering
    \fbox{
    \includegraphics[width=1.0\textwidth]{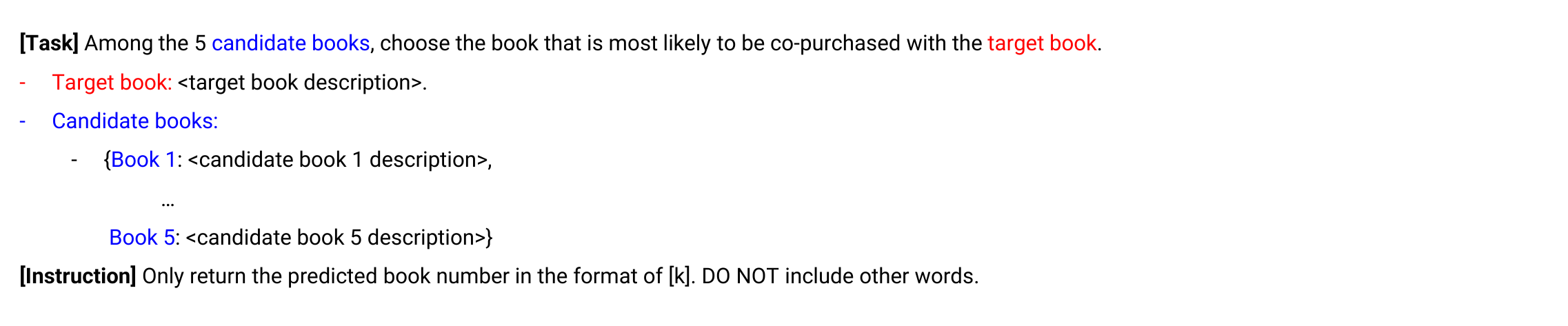}
    }
    \caption{Example edge prediction prompt of \method for co-purchase networks (Book-History dataset).}  
    \label{fig:bookedgeprompt}  
\end{figure*}

\begin{figure*}[t!]
    \centering
    \fbox{
    \includegraphics[width=1.0\textwidth]{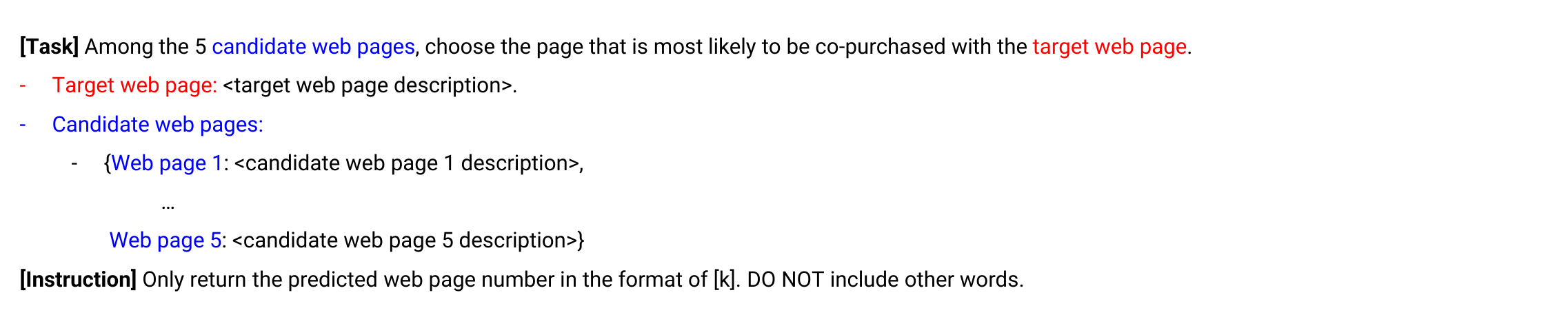}
    }
    \caption{Example edge prediction prompt of \method for hyperlink networks (Wiki-CS dataset).}  
    \label{fig:pageedgeprompt}  
\end{figure*}

\noindent\textbf{Link prediction}
Example link prediction prompts for the OGBN-arXiv dataset (citation network), Book-History (co-purchase), and Wiki-CS (hyperlink network), are provided in Figure~\ref{fig:arxivedgeprompt}, \ref{fig:bookedgeprompt}, and \ref{fig:pageedgeprompt}, respectively.
Specifically, we present the LLM with four randomly sampled nodes and one ground-truth node, prompting it to select the node most likely to be linked to the target node. 
The prompt is tailored to reflect the semantics of the edge type. 
For example, in a co-purchase network, we ask: \textit{`Which book is most likely to be co-purchased with the target book?'}.


\section{Future work}
\label{appendix:futurework}
In this section, we outline potential directions for future work. Several points raised in the Limitation section suggest promising avenues.
Furthermore, \method holds promise for applications beyond the current scope, particularly in graph neural network domains such as anomaly detection~\cite{kim2024rethinking, lee2024slade} and recommendation~\cite{acharya2023llm, gao2022graph}.

\section{License and AI assistant usage}
\label{appendix:aiassistant}
In this appendix section, we discuss (1) the licenses of the artifacts used in this work and (2) our use of an AI assistant, ChatGPT.

\subsection{Licenses}

The licenses of all artifacts used in this work are listed below:
\begin{itemize}[leftmargin=*]
    \item OGBN-arXiv dataset: ODC-BY (\url{https://ogb.stanford.edu/docs/nodeprop/})
    \item Book-History dataset: MIT License (\url{https://github.com/sktsherlock/TAG-Benchmark})
    \item Wiki-CS dataset: MIT License (\url{https://github.com/pmernyei/wiki-cs-dataset})
    \item \texttt{PromptGFM}: CC-By-4.0 (\url{https://arxiv.org/abs/2503.03313})
    \item GPT API: OpenAI permits academic use of the outputs generated by their models (\url{https://openai.com/policies/}).
    \item Claude API: Anthropic permits academic use of the outputs generated by their models (\url{https://www.anthropic.com/legal/commercial-terms}).
\end{itemize}
Note that all permits use for academic purposes.

\subsection{AI Assistant usage}

For this work, we used ChatGPT~\citep{achiam2023gpt} to assist with writing refinement and grammar checking.

\end{document}